\documentclass[10pt,journal,letterpaper,compsoc,twoside]{IEEEtran}

\usepackage{lineno,hyperref}
\usepackage{amsmath}
\usepackage{amssymb}
\usepackage{subfigure}
\usepackage[ruled]{algorithm2e}
\usepackage{tikz}
\usepackage{pgfplots}
\modulolinenumbers[5]

\pgfplotsset{compat=1.14}

\begin{document}

\author{Ri\v{c}ards Marcinkevi\v{c}s, Steven Kelk, Carlo Galuzzi and Berthold Stegemann\IEEEcompsocitemizethanks{\IEEEcompsocthanksitem R.~Marcinkevi\v{c}s, S.~Kelk and C.~Galuzzi are with the Deparment of Data Science and Knowledge and Engineering (DKE), Maastricht University, P.O. Box 616, 6200 MD Maastricht, The Netherlands. B.~Stegemann is with Bakken Research Center (BRC), Medtronic Plc, Maastricht, the Netherlands. S.~Kelk is corresponding author, e-mail: steven.kelk@maastrichtuniversity.nl}}

\title{Discovery of Important Subsequences in Electrocardiogram Beats Using the Nearest Neighbour Algorithm}

\maketitle

\newcommand{\steven}{\textcolor{black}}
\newcommand{\ricardstodo}{\textcolor{red}}

\renewcommand{\thesubfigure}{\alph{subfigure}}
\definecolor{mycolor1}{rgb}{1,0,1}
\definecolor{mycolor2}{rgb}{0,0,1}
\tikzset{mark options = {mark size = 4}}

%\begin{frontmatter}

%\author[brcaddress,dkeaddress]{Ri\v{c}ards Marcinkevi\v{c}s\corref{mycorrespondingauthor}}
%\ead{r.marcinkevics@student.maastrichtuniversity.nl}

%\author{Steven Kelk}

%\author{Carlo Galuzzi}

%\author{Berthold Stegemann}

%\address[brcaddress]{Bakken Research Center (BRC), Medtronic Plc, Maastricht, the Netherlands}
%\address[dkeaddress]{Department of Data Science and Knowledge Engineering (DKE), Maastricht University, Maastricht, the Netherlands}

\begin{abstract}
The classification of time series data is a well-studied problem with numerous practical applications, such as medical diagnosis and speech recognition.  A popular and effective approach is to classify new time series in the same way as their nearest neighbours, whereby proximity is defined using Dynamic Time Warping (DTW) distance, a measure analogous to sequence alignment in bioinformatics. However, practitioners are not only interested in accurate classification, they are also interested in \emph{why} a time series is classified a certain way. To this end, we introduce here the problem of finding a minimum length subsequence of a time series, the removal of which changes the outcome of the classification under the nearest neighbour algorithm with DTW distance. Informally, such a subsequence is expected to be relevant for the classification and can be helpful for practitioners in interpreting the outcome. We describe a simple but optimized implementation for detecting these subsequences and define an accompanying measure to quantify the relevance of every time point in the time series for the classification. In tests on electrocardiogram data we show that the algorithm allows discovery of important subsequences and can be helpful in detecting abnormalities in cardiac rhythms distinguishing sick from healthy patients.

\end{abstract}

%\begin{keyword}
%Electrocardiogram \sep time series classification \sep time series subsequences \sep Dynamic Time Warping \sep nearest neighbour algorithm
%\end{keyword}

%\end{frontmatter}


% to do:
% ORCID
% bio
% foto
% practicioners?
% sec 2.5, learning a classifier
% electrocardiagram (electrocardiogram)?

\section{Introduction} \label{introduction}

Physiological signals are crucial in disease assessment and management. Signals recorded by a digital device, such as a heart monitor, can be represented as time series i.e. measurements performed at regular, discrete time intervals. The automated classification of time series into discrete classes is a natural task. For example, cardiologists might want
to draw on computational analysis of electrocardiogram (ECG) data to determine whether a patient has heart problems. Many different approaches have been proposed in the literature for time series classification \cite{weightedDTW,elasticEnsemble,periodogram,svm,rnn}. Among those the nearest neighbour (NN) algorithm with Dynamic Time Warping (DTW) distance \cite{berndt1994} still outperforms several more recent techniques in terms of the accuracy of classification \cite{bagnall2017}. Essentially this algorithm classifies a newly observed (\emph{unlabelled}) time series by looking at its nearest neighbours in a set of previously classified (\emph{labelled}) time series, and taking a majority vote. DTW distance is used to measure proximity and is similar in spirit to sequence alignment on genomic data (formal definitions will follow). Despite the fact that current research largely focuses on designing \textit{accurate} classifiers, which under experimental conditions typically place unlabelled time series in the correct class, we might not always be interested only in the accuracy.  In particular, we might also wish to know \textit{why} an instance was classified in a particular way. Such questions become especially pertinent in light of new (European) regulations demanding transparency in computer-supported decision-making \cite{goodman2016european}.

Here we propose to answer the \emph{why?} question by studying modifications (i.e. perturbations) of the input sufficient to alter the classification outcome. In particular, leveraging a parsimony perspective, it may be useful to find a \emph{simplest} such modification. 
We formalise this as follows. \textit{Given a labelled dataset of time series $\mathbb{T}$ with a discrete set of classes $Y$
and an unlabelled instance $T$, find a minimum length subsequence of the instance the removal of which changes the outcome of \steven{its}
%the
classification under the nearest neighbour algorithm with DTW distance}.
%\textcolor{blue}{Is the whole article only about $|Y|=2$ or does your method support more classes?}
Observe that, if $|T|=n$, the number of subsequences of the time series is $2^n$. Therefore, a na\"{i}ve search through the space of all candidate solutions is impractical. The problem can be simplified by considering only \emph{contiguous} subsequences of which there are $\Theta(n^2)$. In this article we consider only this simplified version of the problem, but as we shall demonstrate this is still powerful enough to detect interesting features in real-world data.

\begin{figure}
\centering
\includegraphics[scale=0.14]{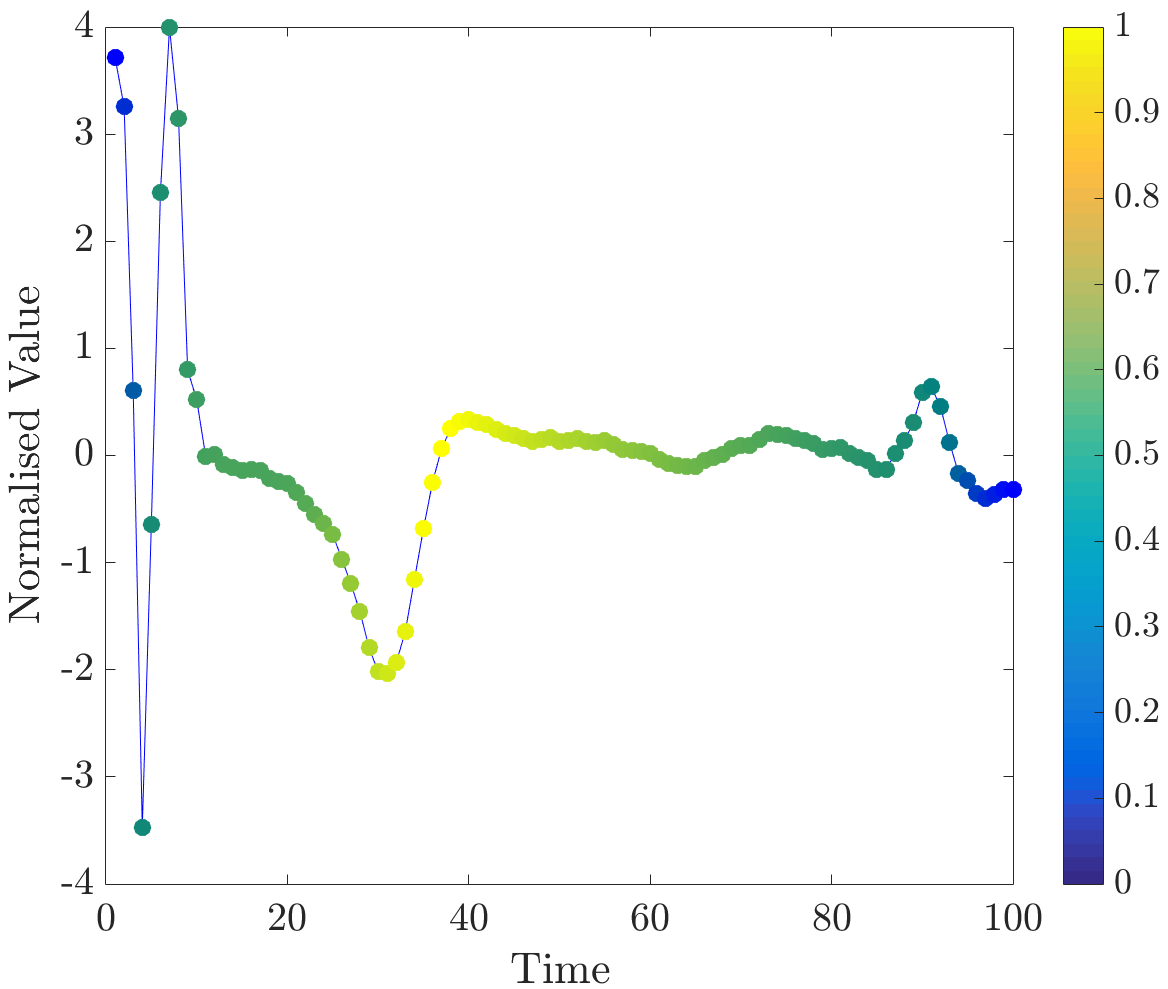}
\caption{An ECG beat (from the V3 channel) of a patient diagnosed with ARVD, a rare cardiac disease. One of the 
manifestations of ARVD in ECGs is the negative polarity of T waves (compared to healthy subjects) \cite{marcinkevics2017}. The points in the plot are coloured according to their importance for the 
classification, as computed by our method (yellow being the most important and blue the least).
As can be seen, the algorithm successfully grasps the importance of the T wave in this case \steven{i.e. the downwards-pointing segment
that peaks at -2. In healthy patients this segment points upwards.}}
\label{fig:heartintuition}
\end{figure}

The structure of this article is as follows. In Section \ref{sec:prelim} we provide background information on time series data, the nearest neighbour algorithm, DTW distance and also some basics on interpreting electrocardiogram (ECG) data (which forms the core of our experimental study). We also explain how our approach differs from other approaches in the
literature.

 In Section \ref{methods} we describe our approach for solving the problem, \steven{which consists of the natural subsequence search algorithm augmented with a number of enhancements to reduce the running time, such as efficient re-use of the DTW dynamic programming table and pruning.}
%optimizing the dynamic programming approach used to compute DTW distance.
We also introduce a measure for assessing the relevance of points in the time series data: informally, how frequently they occur in minimum modifying subsequences. Section \ref{results} presents and interprets the results of our experiments on ECG data. The intuition is that minimum modifying subsequences will potentially contain ECG waveform segments and points which are relevant to the underlying diagnostic problem. This in turn may help to interpret the classification result for specialists and even to discover new knowledge about the morphology of the ECG under certain diseases. The results are
%very
encouraging and support this intuition; \steven{see Figure \ref{fig:heartintuition} which is reproduced from our experimental section. Another set of experiments we conduct demonstrates that our method is not only useful for interpretation and discovery of important subsequences, but also (when used `backwards') as an instrument for detecting and locating \emph{known} features.} 
Finally, in Section \ref{concl.} we reflect on our work and consider solution perspectives for the non-contiguous version of the problem.

%the number of which is $\frac{1}{2}n(n+1)$ \cite{elzinga2008}.

%The method discussed and tested further considers only contiguous subsequences.

%but even then the running time of the na\"{i}ve algorithm is compounded by the time needed to repeatedly run the nearest neighbour search with the DTW distance through the set of labelled instances. To conclude, an efficient search algorithm needs to be proposed to address the aforementioned problem for large datasets with long time series.

\section{Preliminaries}
\label{sec:prelim}

\subsection{Time Series Data}

%For the sake of convenience, herein we introduce some notation for this form of data, which is used in further sections. 
Electrocardiograms and many other %forms of periodically measured
signals can naturally be represented by \emph{time series}.
A time series $T$ with $n$ points, equally spaced in time, is described by a sequence of values $\langle T_1, T_2, ..., T_n\rangle$, where $T_i$ refers to the value measured at the $i$-th time point. Based on this simple representation, signals can be compared, clustered or classified \cite{ding2008}. In particular, comparison can be performed using some (dis\nobreakdash-)similarity measure, for example, Euclidean distance. For a good overview of time series data mining, see \cite{ding2008} and \cite{esling2012}.

\subsection{The Nearest Neighbour Algorithm and Search}

%Nearest neighbour classification is at the core of the considered problem, because the target is to alter the outcome of this algorithm.
The \emph{$k$-nearest neighbour ($k$-NN) algorithm} is perhaps the simplest instance-based classification technique and has been well-studied in the machine learning literature \cite{CoverHart1967,wilson1972,Friedman1975}. Given a dataset of labelled training instances and an unobserved instance to be classified, this algorithm returns the label that is most common among $k$ training instances closest to the unobserved instance, according to some distance function \steven{(and breaking ties in some systematic way)} \cite{mitchell1997}. \steven{In this article, to clarify the exposition, we focus on the case} $k=1$ and assume only two classes; however, all of the methods presented can be easily generalised to non-binary classification with $k>1$. 

The NN algorithm includes solving an optimisation problem known as \emph{nearest neighbour searching} \cite{clarkson2006}, formally defined in the following way. Given set $\mathbb{S}$ of instances, distance function $d$ and query instance $Q$, we need to find instance $S\in \mathbb{S}$, which will minimise $d(S,Q)$. Depending on the distance
function various refinements can be made to accelerate identification of this minimum \cite{clarkson2006,pauleve2010}.
%In general, many enhancements can be introduced to the nearest neighbour searching  under certain assumptions about the distance function.

%The na\"{i}ve solution is to linearly search through all elements in $\mathbb{S}$. The distance function can be sometimes expensive to evaluate, therefore, `cheaper' lower bounding functions can be used to enhance the linear search. Moreover, under certain assumptions about the function, data structures for efficient NN searching can be constructed. In particular, if $d$ is a metric, Orchard's algorithm, the Approximating and Eliminating Search Algorithm (AESA) and metric trees can be applied to solve the problem \cite{clarkson2006}. Approximate solutions are addressed in the literature as well. An example of a viable approximation technique for high-dimensional data is Locality-sensitive Hashing (LSH) \cite{pauleve2010}.

\subsection{Dynamic Time Warping}
A viable distance function for the NN algorithm is \emph{Dynamic Time Warping (DTW)} distance. This a distance measure for time series data, used in a wide range of domains \cite{rakthanmanon2012}. It is evaluated based on a least `cost' alignment between two time series.
%\steven{computed} by a dynamic programming (DP) algorithm.
Although the details differ it is analogous to the task of pairwise sequence alignment of DNA sequences encountered in bioinformatics.

The DTW distance between two time series $S$ and $T$, $n$ and $m$ points long, respectively, can be defined recursively as follows \cite{kim2001}:
\[
\begin{split}
d_{DTW}(S,\langle\rangle)={ }&\infty,
\\
d_{DTW}(\langle\rangle,\langle\rangle)={ }&0,
\\
d_{DTW}(S,T)={ }&\delta(S_n,T_m) + \min\{d_{DTW}(S_{1:n-1},T),
\\
&d_{DTW}(S,T_{1:m-1}),
\\
&d_{DTW}(S_{1:n-1},T_{1:m-1})\},
\end{split}
\]
where $\langle\rangle$ is the empty series; $\delta(\cdot)$ denotes a local distance function (we assume $\delta(S_i,T_j)=|S_i-T_j|$); and $S_{i:j}$ stands for the contiguous subsequence of time series $S$ beginning in the $i$-th time point and ending with the $j$-th. \steven{Note that $n$ and $m$ do not need to be equal.}

Based on the recurrence relation above DTW distance can be computed by dynamic programming. This aligns the two sequences by finding an optimal warping path through the $n\times m$ grid of distances \cite{berndt1994, ahmed2017}. 
The running time complexity of the standard implementation is $O(nm)$ \cite{kim2001, muller2007}.

The optimal warping path satisfies monotonicity and continuity constraints \cite{berndt1994}, i.e. aligned points are ordered monotonically w.r.t. time and the steps of the path are confined to neighbouring points in the grid. Additionally, a warping window constraint can be imposed \cite{berndt1994} by allowing warping paths only within a certain window.

In order to avoid computationally expensive evaluations of the DTW distance during NN search,
%often `cheap'
quickly computable lower bounding functions are often utilized. \steven{If the lower bound is already too large the time series
cannot possibly be one of the $k$ nearest neighbours and can be immediately excluded.} Several lower bounds have been proposed in the literature \cite{kim2001, keogh2005, yi1998}. It is also important to mention that fast (linear in time) approximations to the DTW distance exist. For instance, the FastDTW algorithm approximates the distance by computing the alignment between down-sampled time series \cite{salvador2007}.

\subsection{Electrocardiograms}

We have tested our method on electrocardiogram datasets. The surface electrocardiogram (ECG or EKG) quantifies the electrical potential between points on the body \cite{karel2009}. It is recorded by placing electrodes (leads) on the body in various sites \cite{karel2009}.
%Nowadays,
This technique is widely used for the measurement of heart activity and can be utilized in the diagnosis and management of various pathologies.

\begin{figure}[ht]
\begin{center}
\includegraphics[width=4.0cm]{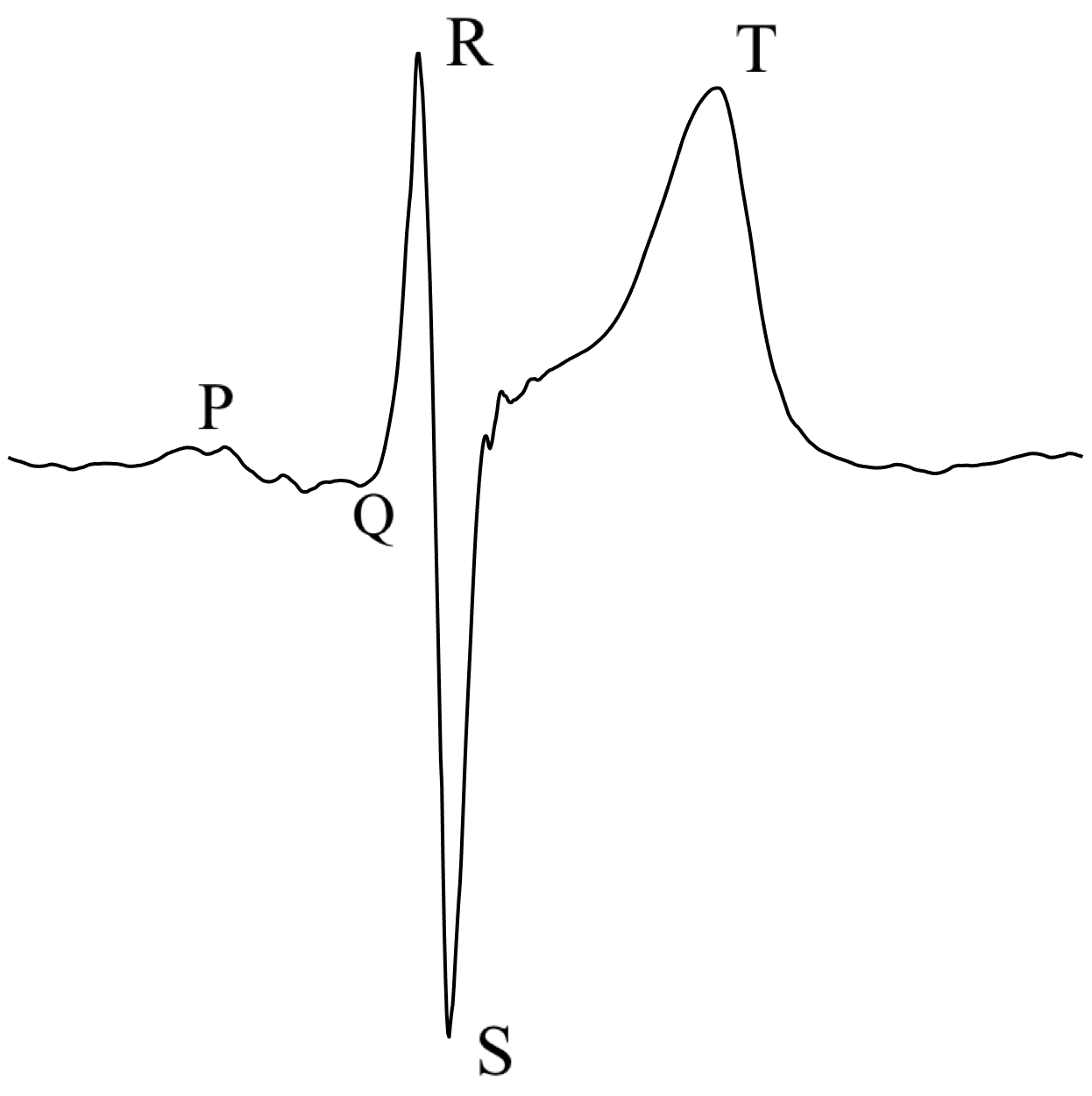}
\caption{An ECG beat and its characteristic landmarks \steven{in a healthy patient}. $QRS$ complex, $P$ and $T$ waves are emphasised.\label{fig:beat}}
\end{center}
\end{figure}

An ECG signal can be characterised by various relevant points, segments and waves. Figure \ref{fig:beat} depicts a beat from an ECG of a healthy person; $T$ and $P$ waves, as well as points $Q$, $R$ and $S$ are marked in the plot. All of these landmarks have a physiological meaning in terms of the cardiac cycle \cite{karel2009}.
It is important to note that the morphology of the beats can vary dependending on the electrode position and on the disorders that the heart suffers from \cite{karel2009}.

\subsection{Related work}

%The temporally-ordered nature of time series data requires
The classification of time series requires a tailored approach due to the temporal ordering of its datapoints.
%Due to the temporal ordering of datapoints in time series, the classification of  Time series classification is a specific problem because the features of data points are ordered temporally.
Many algorithms have been proposed to classify time series data. Broadly speaking there are two major approaches to the problem: feature- and distance-based \cite{Susto2018}. The former amounts to extracting a set of features from raw time series prior to
%\steven{training} ?
training
a classifier \cite{Susto2018}, examples of simple features are averages and standard deviations of time series intervals. A feature-based approach to classification is discussed in \cite{svm}, where time series are classified by a support vector machine (SVM) trained on predicate features extracted from intervals. The latter approach, on the other hand, quantifies dissimilarities between time series (raw or transformed) and then usually applies an instance-based classification algorithm \cite{Susto2018}. For example, the $k$-nearest neighbours algorithm with weighted DTW distance, proposed in \cite{weightedDTW}, is a case of distance-based classification. Most of the approaches, distance- or feature-based, provide little explanatory power. While literature about training accurate classifiers and understanding them is abundant, the problem of interpreting \emph{individual} classification outcomes and discovering important intervals within time series is addressed less frequently. 

To the best of our knowledge our approach has not been described in the literature before. However, two primitives from the time series data mining literature exhibit some (superficial) similarity to our approach.  The problem of finding a time series \emph{discord} \cite{keogh2005disc}, the most `unusual' subsequence, is relevant, because it targets the discovery of important time series segments. A discord is formally defined as a contiguous subsequence of a time series, which has a maximum distance to its closest non-self match \steven{i.e.} another subsequence that does not overlap with it. Discords were demonstrated to be useful in anomaly detection in various sensor time series, such as ECGs. Other relevant time series primitives are \emph{shapelets} \cite{ye2009}. A shapelet can be intuitively viewed as a time series subsequence that is identified as being representative of \steven{an entire} class in a two-class dataset. For the formal definition of shapelets the reader is referred to \cite{ye2009}. Shapelets can be used for a fast and interpretable classification.

Both of these primitives involve the analysis of subsequences in time series data and in that sense resemble our approach. However, neither involves the NN algorithm or the DTW distance (explicitly). Moreover, these primitives are
not intended for interpreting \emph{classification} outcomes at the level of \textit{individual} time series. Discord is useful for
detecting anomalies within a single time series (i.e. it does not consider other time series) and shapelet analysis can be used to identify general characteristics of a class - but it does not act at the level of an individual time series. For this reason our approach is not directly comparable to these earlier approaches and we do not consider them in our experiments.

\section{Method} \label{methods}

%The problem of finding a minimum \steven{length} subsequence of the given time series, the removal of which changes the outcome of classification can be relaxed by considering only contiguous subsequences. The latter assumption reduces the size of the solution space. In the remainder of this section, we explain the procedure for finding a minimum contiguous subsequence. In addition, at the end of this section, we describe a measure to quantify the \textit{relevance} of every time point in the time series for the classification. This measure can be applied in interpreting the outcome of the time series classification.

As mentioned before, there are only $\Theta(n^2)$ many \emph{contiguous} subsequences of a time series. Therefore, it is possible to search through all of them exhaustively \steven{for reasonable-sized $n$}. Algorithm \ref{alg:contsearch} contains the \steven{conceptual} pseudocode for the \textit{na\"{i}ve} search, \steven{for $k=1$}, which iteratively considers subsequences of increasing lengths. 
\begin{algorithm}[ht]
    \SetAlgoLined
    \SetKwInOut{Input}{Input}
    \SetKwInOut{Output}{Output}
    \Input{Labelled time series dataset $\mathbb{T}=\{\langle S_1,y_1\rangle,\langle S_2,y_2\rangle,...,\langle S_k,y_k\rangle\}$ and unobserved time series $T=\langle T_1,T_2,...,T_n\rangle$.}
    \Output{A minimum length contiguous subsequence of $T$ the removal of which changes the classification outcome.}
    \bigskip
    let $\textrm{class}\left[\langle X,y\rangle\right]=y$\\
    \bigskip
    \tcp{Initial classification}
    $c\gets\textrm{class}\left[\arg\min_{\langle S_i,y_i\rangle\in\mathbb{T}}d_{DTW}(S_i,T)\right]$\;
    \bigskip
    \tcp{Subsequences of length $i$ at position $j$}
    \For{$i=1$ to $n-1$} {
        \For{$j=2$ to $n-i$} {
            \tcp{Modified time series}
            $T'\gets T_{1:j-1}\Vert T_{j+i:n}$\;
            $c'\gets\textrm{class}\left[\arg\min_{\langle S_i,y_i\rangle\in\mathbb{T}}d_{DTW}(S_i,T')\right]$\;
            \If{$c'\neq c$} {
                \Return $T_{j:j+i-1}$\;
            }
        }
    }
    \caption{Contiguous Subsequence Search.}\label{alg:contsearch}
\end{algorithm}
Herein, operator ``$||$'' denotes concatenation of sequences. Note, that in this search subsequences containing end-points of $T$ are not considered; we make the assumption, for the sake of convenience, that the end-points may not be removed. This algorithm is merely a na\"{i}ve search procedure and can be further optimised.

As can be seen, the essence of the search is simply to consider all subsequences of increasing length and to perform NN classification each time. It is difficult to construct efficient data structures for NN searching under the DTW distance, because it is non-metric and the assumptions made by many other algorithms (in particular: that the triangle inequality holds) are not valid here \cite{clarkson2006, pauleve2010}. Moreover, \steven{many} tight lower bounding functions on the distance proposed in the literature are also not applicable, because they assume that the two time aligned series are of the same length \cite{kim2001, keogh2005, yi1998}. \steven{Due to the deletion of subsequences this assumption also does not hold in our case.} \steven{For these reasons we have elected to use the standard, linear NN search.} Nevertheless there are several practical improvements which can be applied; many of them are \steven{tailored to our specific, DTW-based deletion approach and are thus not implied by more general techniques for NN search optimization.} 
%and are not directly relevant to general NN searching (and .
Namely, we enhanced the procedure by
\begin{itemize}
\item early abandoning \cite{rakthanmanon2012} and pruning \cite{silva2016} of DTW distance computations;
\item re-using old DTW alignments in new distance computations and lower bounds;
\item repairing the DTW distance function \cite{clarkson2006} to make it satisfy the triangle inequality, from which lower bounds can be derived.
\end{itemize}

% NEED TO ADD A PIECE THEORY HERE!!!!!!!!!!!!!!!!!!!!!!!!!!!!!!
Early abandoning and pruning distance computations as well as repairing distance functions has been discussed in the literature before, and we refer the reader to \cite{clarkson2006, rakthanmanon2012, silva2016} for more details. Nevertheless, re-using DTW alignments is novel to this problem and is explained further.

%throughout computations is only relevant to this problem 

When initially classifying an unobserved time series $T$, DTW alignments between $T$ and training instances from dataset $\mathbb{T}$ have to be computed. These alignments can be re-used further. We demonstrate that lower bounds on distances between a modified time series and training instances can be inferred from these alignments. Moreover, they can be re-used when computing new alignments, if needed.

\begin{figure}[ht]
\begin{center}
\includegraphics[width=5.0cm]{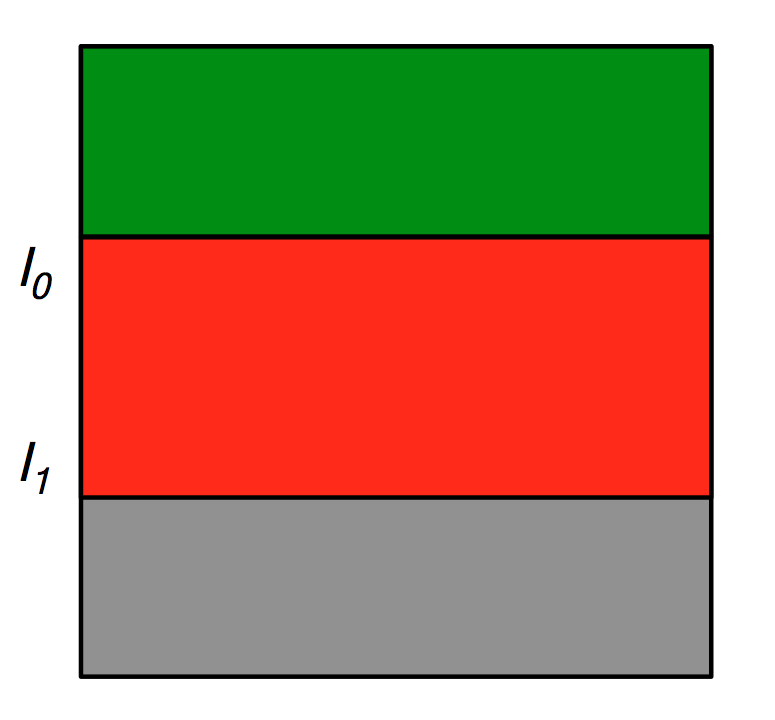}
\caption{A dynamic programming table for aligning two sequences. If a subsequence from row $l_0$ to $l_1$ is removed, then row $l_0-1$ contains a lower bound on the DTW distance; and the green part of the table can be re-used in the further alignments.\label{fig:reusing}}
\end{center}
\end{figure}

Let $T'$ denote a times series that resulted from a modification of $T$. $T'$ was constructed by removing a contiguous subsequence from $T$, such that the left-most point of this subsequence is at index $l_0$ and the right-most is at index $l_1$. Consider the DTW alignments between $T$ and some training instance $S\in \mathbb{T}$ and between $T'$ and $S$. By construction, dynamic programming tables for these alignments, $D$ and $D'$, respectively, are identical from row $0$ to row $l_0-1$ inclusively. This is schematically illustrated in Figure \ref{fig:reusing}. 
Based on this invariance, for every time series in the dataset, a lower bound on its distance to the modified time series $T'$ can be derived from its initial alignment with $T$. If both $l_0$ and $l_1$ are placed close to the left end-point of $T$, then this lower bound will not be very tight. In that case a tighter lower bound may be derived in the same manner from the alignment between $T^R$ and $S^R$, {which are the reversals of $T$ and $S$ respectively.

What is the utility of the aforementioned lower bounds? They can be used to prune the \steven{standard, linear} nearest neighbour search
%\footnote{\steven{This is the variant of NN search that we have used in this article. That is, where we simply compare the unobserved instance to each of the labelled datapoints in turn. Data-structure enhancements such as kd-trees and hashing can sometimes be deployed to speed up NN search but these have so far proven to have limited efficacy when DTW is the underlying distance.}}
and avoid expensive DTW distance evaluations. In particular, given a modified time series $T'$, some $S$ from the training set and upper bound $\tau$ on the distance to the NN of $T'$, $S$ can be pruned in the NN search if \(\tau<LB_{DTW}(T',S),\) where $LB_{DTW}$ is a lower bound on $d_{DTW}$.

Apart from computing lower bounds on distances, the initial alignments can be stored in memory entirely, to re-use the DP tables for modified time series. In general, if a subsequence from $l_0$ to $l_1$ is removed from $T$, then the computations of the alignment between $T'$ and any time series in the training set can be started from row $l_0$ based on row $l_0-1$ of the initial alignment between $T$ and the training instance. In practice this can speed up the algorithm considerably at the cost of increasing the space complexity, because complete dynamic programming tables have to be stored.

\subsection{Relevance of Time Points} \label{relevance}

A minimum length subsequence the removal of which changes the outcome of classification is expected to contain points of the time series that are relevant to the considered classification task. However, we observed that the minimum length subsequence alone often is not sufficiently interpretative. Therefore, we propose a measure that, intuitively, is expected to quantify the relevance of every point in the time series for the classification. 

For an unobserved time series $T$, let $\mathcal{S}(i)$ denote the set of contiguous subsequences, the removal of which changes the outcome of classification of $T$ and which contain the $i$-th point of $T$. For every point $i$, \emph{relevance} $r$ \mbox{is given by} \[r(i)=\sum_{Q\in\mathcal{S}(i)}\frac{1}{|Q|}.\] The relevance value is thus simply the sum of the inverted lengths of subsequences in $\mathcal{S}(i)$. Therefore, this measure promotes the points which are contained within many subsequences and, especially, the points which are contained within many \textit{short} subsequences.  

Note, that Algorithm \ref{alg:contsearch} can be easily adjusted to calculate the relevance vector for the input time series. Computing relevance values requires at least as many operations as finding the minimum length subsequence. Moreover, non-minimum length subsequences have to be found as well. The problem can be relaxed by considering only those subsequences which begin at every $j$-th time point or by considering subsequences shorter than a given length; this results in a faster approximation of relevance values.

The relevance measure is expected to provide a visual and informative way for interpreting the classification of individual time series, given some dataset. It can be used to identify time series segments particularly relevant to the considered classification problem.

\section{Experiments and Results} \label{results}

In order to verify the designed algorithm and its performance, several experiments were conducted. In this section, we describe the datasets that were used in testing, how the data was pre-processed and, finally, the experiments themselves and their results. \steven{Our first experiment in Section \ref{subsec:scalability} studies the scalability of our method for computing relevance values. In our second experiment, in Section \ref{subsec:interpret}, we use our method to \emph{interpret} classification (through discovery of important subsequences). In the third experiment, Section \ref{subsec:detect}, we use our method `backwards' to  \emph{detect} subsequence types that we have identified a priori as being of interest.}
%Two types of experiments were conducted; namely, to assess the informativeness of relevance values for different classification tasks and to study the scalability of the algorithm for computing relevance values. 
For all experiments a computer with 2.3GHz quad-core Intel Core i7 CPU and 8Gb memory was used. 

\subsection{Datasets and Processing} \label{data}
%\ricardstodo{Are all these datasets used in the experiments? If not, perhaps leave the unused ones out?}
We used several ECG datasets. In particular, two most common classes were taken from the training set of ECG5000 beat dataset, available at the UCR Time Series Classification Archive \cite{ucrarchive}. Other datasets, obtained from this archive, were CinC\textunderscore ECG\textunderscore torso testing set with several classes of ECG beats and two-class ECG500. We also constructed a set of beats from various ECGs available in the PhysioBank databases \cite{physionet}. Finally, ECG beats from the small dataset of normal subjects and patients diagnosed with arrhythmogenic right ventricular dysplasia (ARVD) \cite{marcinkevics2017}, kindly provided by Leeds General Infirmary, were used as well. 

Prior to running tests, some ECGs were denoised in the MATLAB environment using \texttt{wden} function for automatic signal denoising with wavelets, db4 (Daubechies family) wavelet was applied. After that, if needed, the time series were normalised and down-sampled. For some of the experiments, $QRS$ complexes and $T$ waves were labelled manually.

Finally, relevance values were computed for the chosen unobserved time series by the designed algorithm, implemented in Java alongside some important subroutines for (amongst others) the DTW and the NN algorithms.

\subsection{Algorithm Scalability}
\label{subsec:scalability}

An experiment to test the scalability of the implemented procedure was conducted. The running times for computing relevance values for each time point in unobserved time series of varying lengths ($n$) were measured. Measurements were performed for three datasets of different sizes, namely, the set of ARVD beats, ECG500 and ECG5000. 
%($k$)
Figure \ref{fig:runTimes} contains the plot of running times versus different time series lengths for sets of 16 (\ref{fig:runTime_1}), 100 (\ref{fig:runTime_2}) and 500 (\ref{fig:runTime_3}) instances. Relevance values were computed by different versions of the implemented algorithm (see figure legends for details).
\begin{figure}
\centering
  \subfigure[\label{fig:runTime_1}]{\includegraphics[scale=0.171]{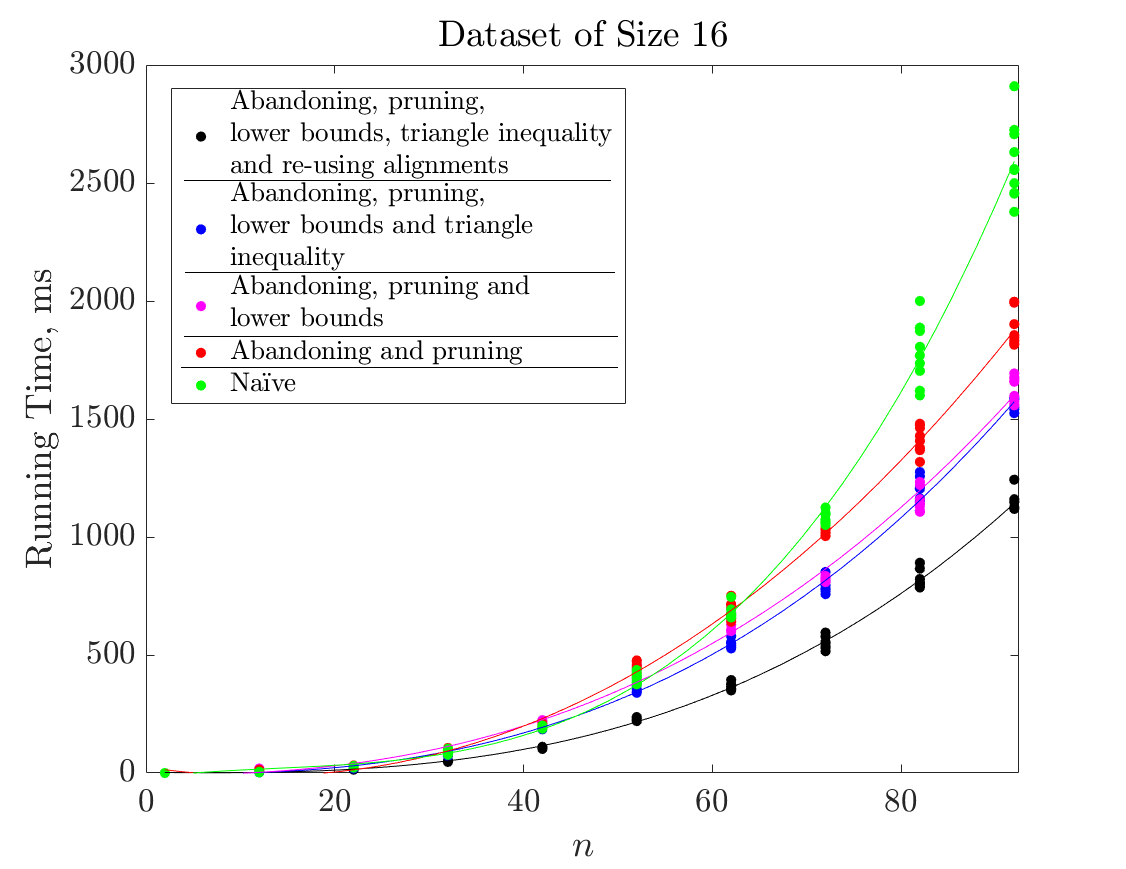}}\hfill
  \subfigure[\label{fig:runTime_2}]{\includegraphics[scale=0.171]{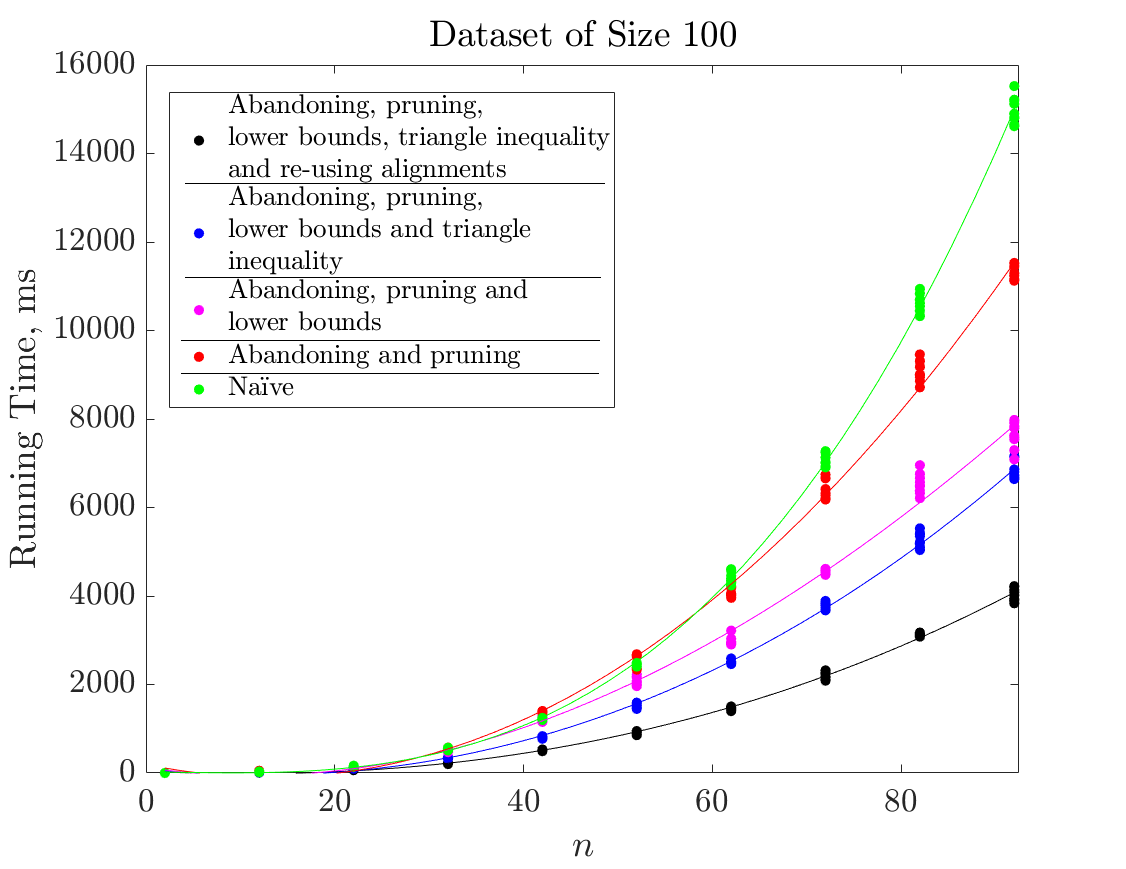}}\hfill
  \subfigure[\label{fig:runTime_3}]{\includegraphics[scale=0.171]{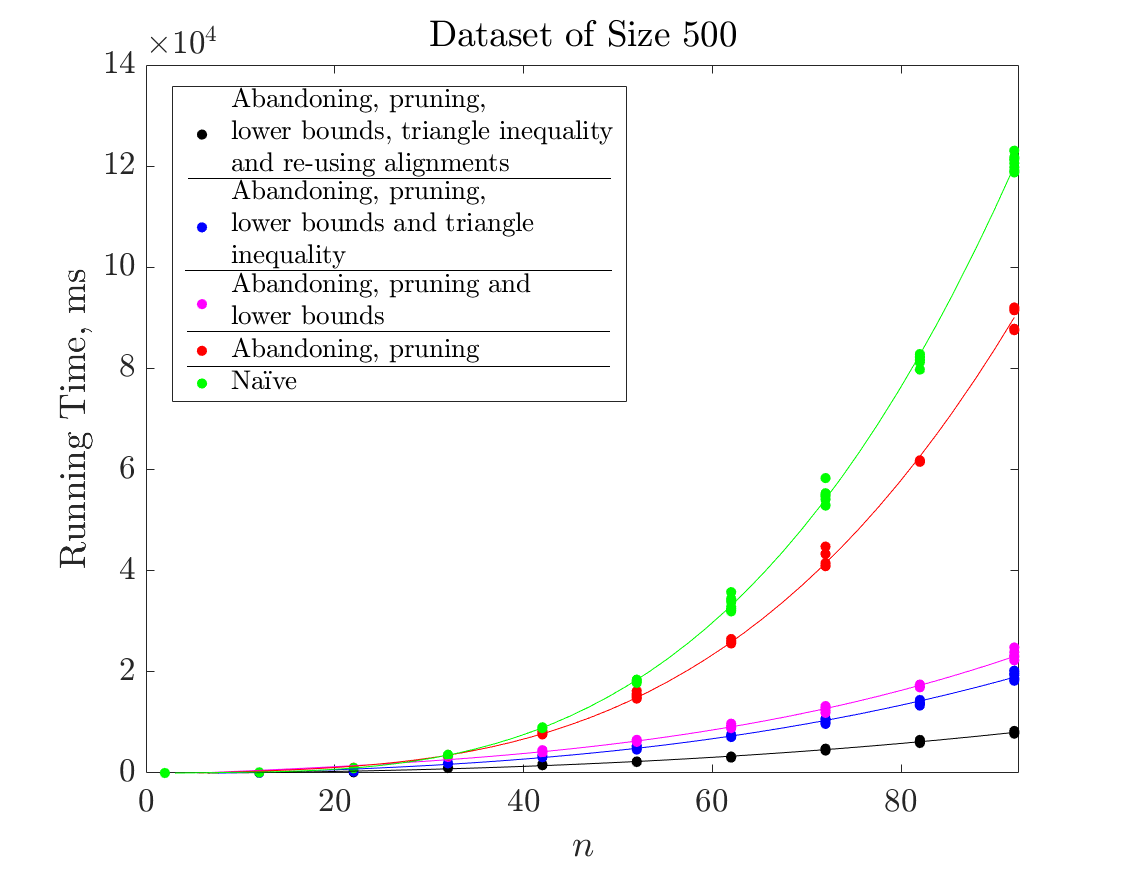}}
\caption{Running times of computing relevance values for different lengths of time series ($n$) %$T$, $n$,
and datasets of different sizes. \label{fig:runTimes}}
\end{figure}
%Cubic polynomials were fitted to the running times. Least squares degree 3 polynomials appear to be suitable; this agrees with the theoretical expectations.

%The results show that, the na\"{i}ve search procedure had the worst performance. For computing relevance values, the enhanced version of the algorithm is more efficient than the na\"{i}ve search. We demonstrated that implemented enhancements,

The implemented enhancements (abandoning and pruning the DTW, inferring lower bounds on distances, re-using alignments \steven{and repairing the triangle inequality}) achieve significant speedups for inputs of larger sizes. Thus, for example, given a dataset of 500 instances, the most efficient version computes relevance values for a time series with 80 points in approximately the same time as the na\"{i}ve version for a time series with 40 points. \steven{In the experiments that follow we used the most efficient version of our algorithm i.e. where all enhancements are activated.}

\subsection{Interpreting ECG Classification}
\label{subsec:interpret}
%\steven{We should state here explicitly which version of the algorithm was used and (somewhere earlier probably) which
%computer was used. It was probably your laptop, right? Some specs are useful here. Also, all experiments are conducted
%with $1-NN$ right?}
Since the motivation for designing the described algorithm was the interpretation of ECG classification outcomes, we performed experiments wherein relevance values were computed for unobserved time series in two different classification tasks. The distributions of relevance values in several instances were inspected visually and interpreted.

\subsubsection{Arrhythmogenic Right Ventricular Dysplasia (ARVD)}

One of the studied classification tasks is \steven{patient classification for}
ARVD, a rare cardiac disease \cite{marcus1982}.
%, patient classification.
For this purpose, a small stratified set of 16 beats from normal and abnormal (ARVD) subjects was used. Denoised ECG beats are plotted in Figure \ref{fig:ARVDdata}.
\begin{figure}[ht]
\begin{center}
\includegraphics[width=6.5cm]{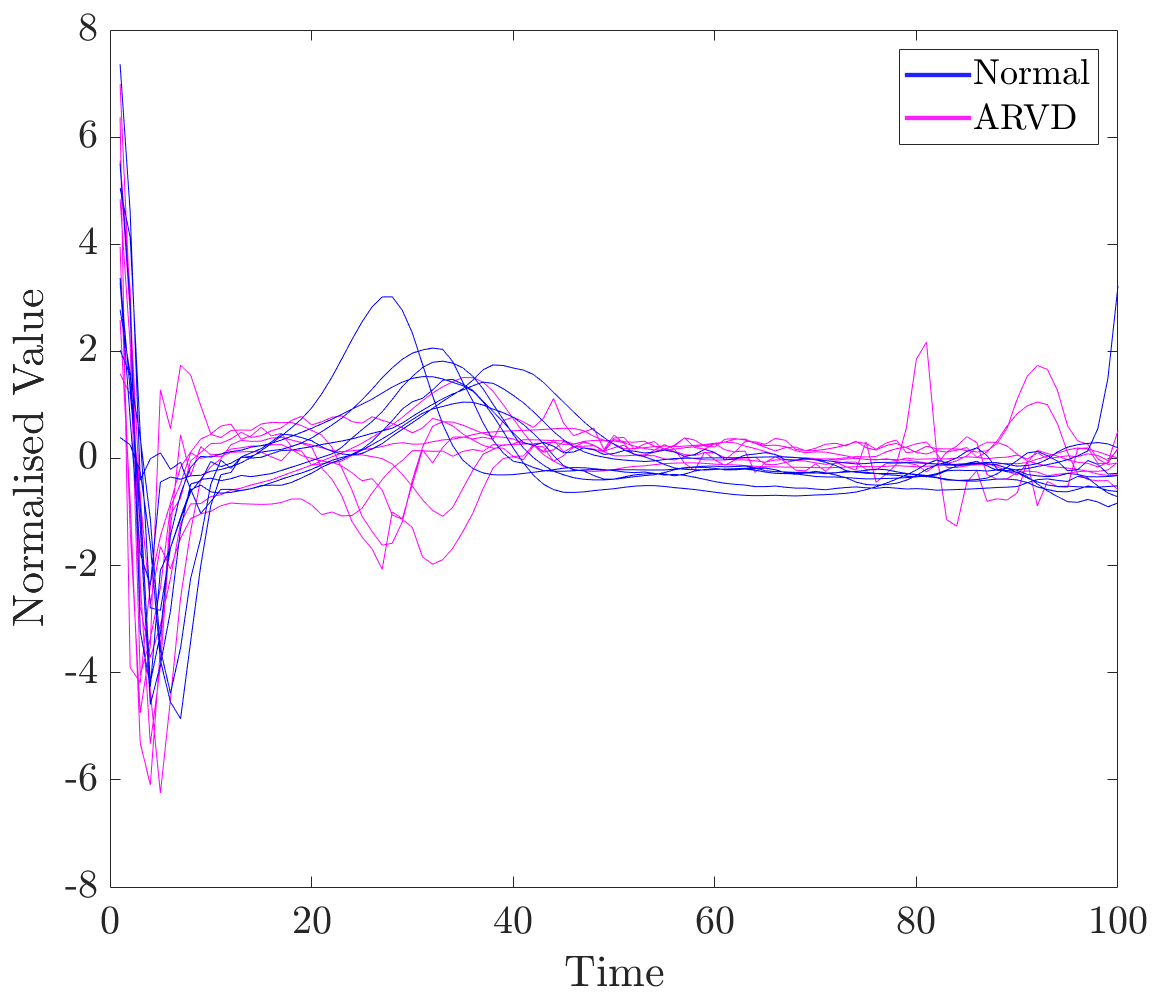}
\caption{ECG beats of normal and ARVD subjects.\label{fig:ARVDdata}}
\end{center}
\end{figure}
According to the description given by a cardiologist \cite{marcinkevics2017}, ``in the ARVD  patients the $T$ wave polarity is more often negative and the $S$ wave of the $QRS$ is more delayed and pronounced, [...] the visualised $P$ wave is larger in  amplitude''. $T$ wave inversion, $QRS$ widening and $P$ wave amplification can be seen in Figure \ref{fig:ARVDdata}. The goal of this experiment is to investigate if relevance values computed for individual time series, when classifying them, could be used to discover important segments mentioned above.

Relevance values were computed for instances that \textit{were classified correctly} by the NN algorithm. \steven{Specifically, the 16 beats in the dataset have already been labelled, so the correct classification for each beat (healthy or abnormal) is already known. A single beat is removed and, if the NN algorithm classifies it correctly, our algorithm is applied.}
%(an instance for which the computations were %performed was removed from the training set). 
Relevance values of all time series were inspected visually. For abnormal beats, high relevance values were often concentrated around $S$ waves (see Figure \ref{fig:abnS}) and sometimes around $T$ (see Figure \ref{fig:abnT}) and $P$ (see Figure \ref{fig:abnP}) waves. 
\begin{figure}
  \centering
  \subfigure[\label{fig:abnS}]{\includegraphics[scale=0.11]{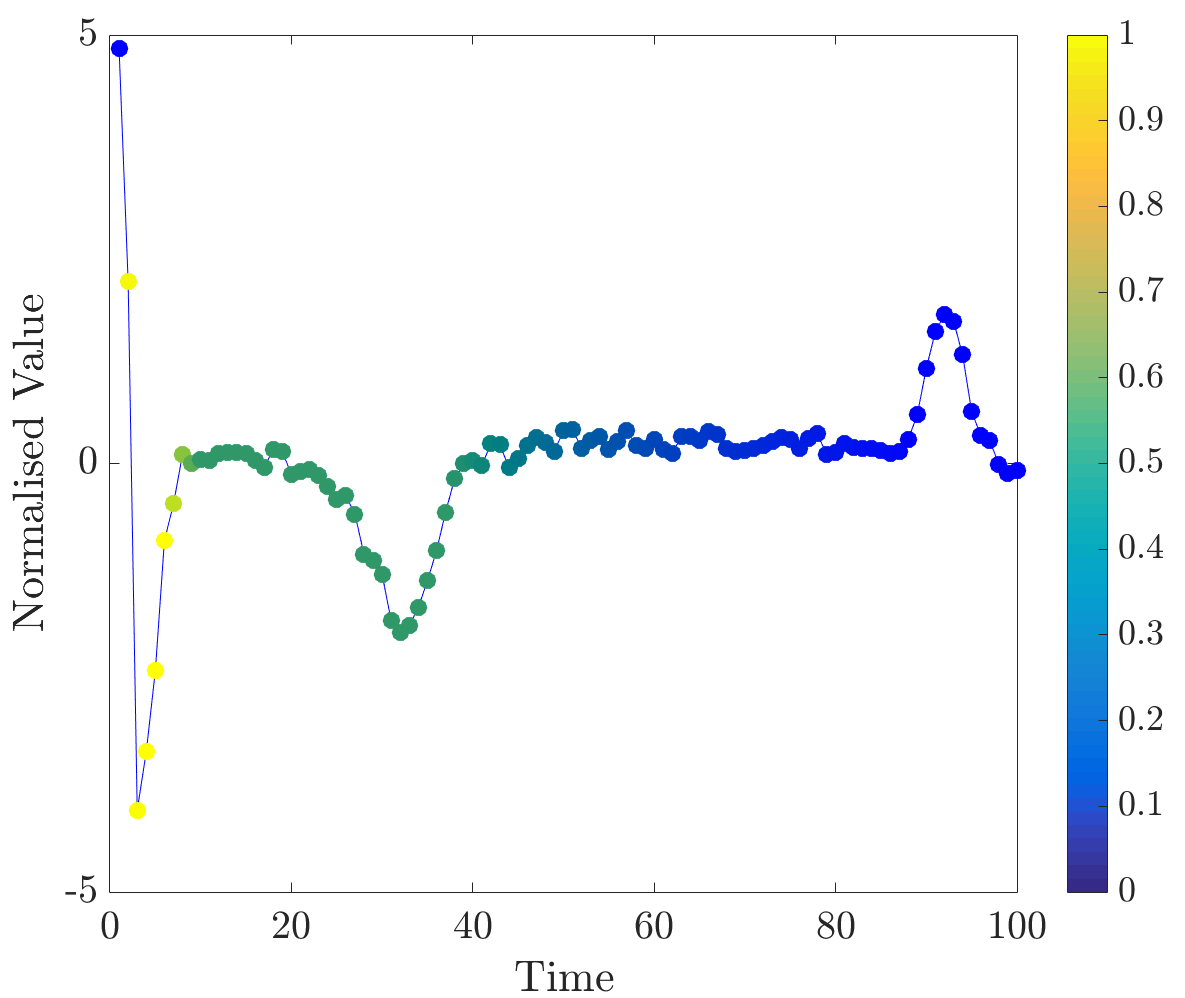}}\quad
  \subfigure[\label{fig:abnT}]{\includegraphics[scale=0.11]{images/arvd_abnormal_Twave.png}}\quad
  \subfigure[\label{fig:abnP}]{\includegraphics[scale=0.11]{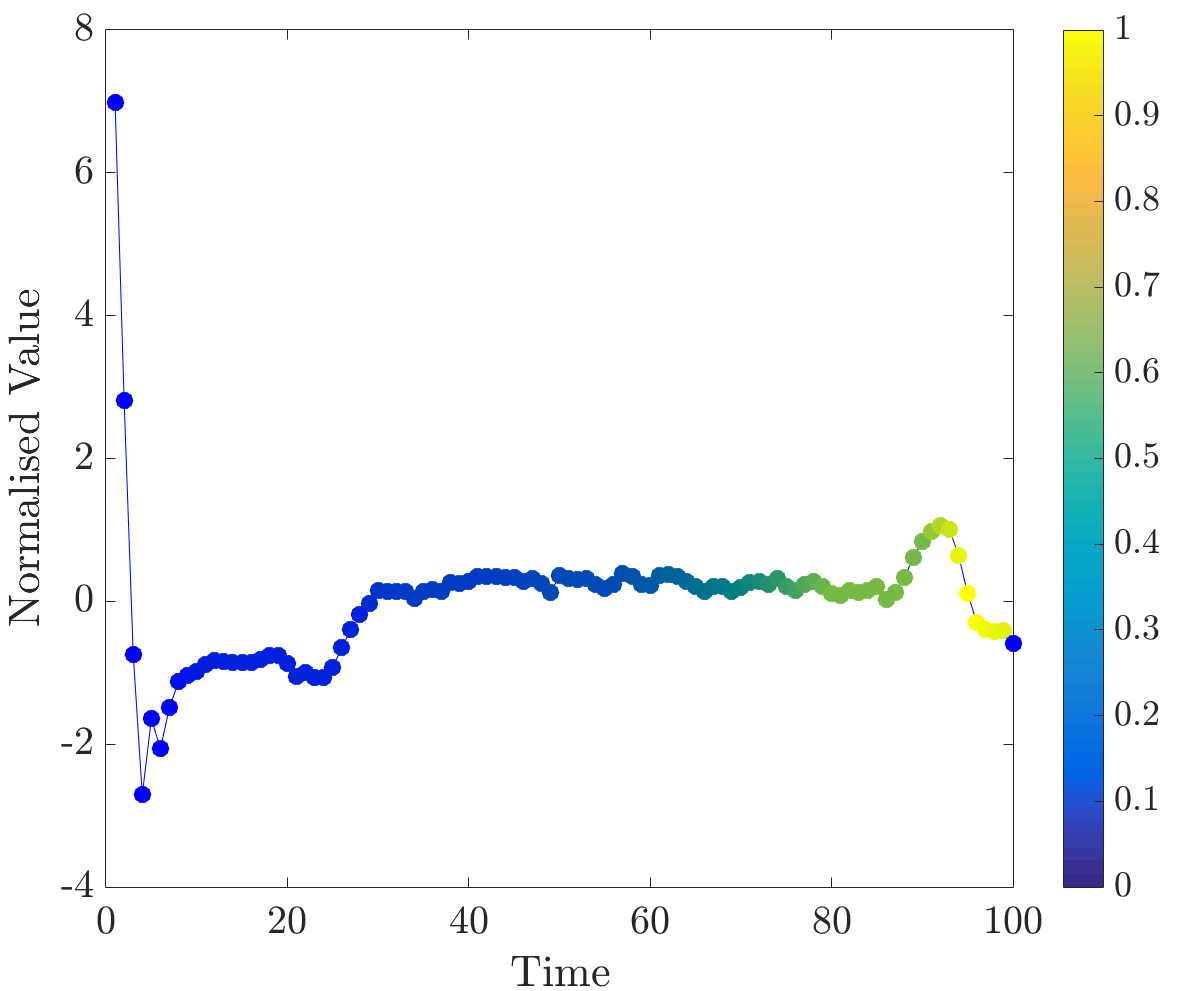}}\quad
  \subfigure[\label{fig:abnStr}]{\includegraphics[scale=0.11]{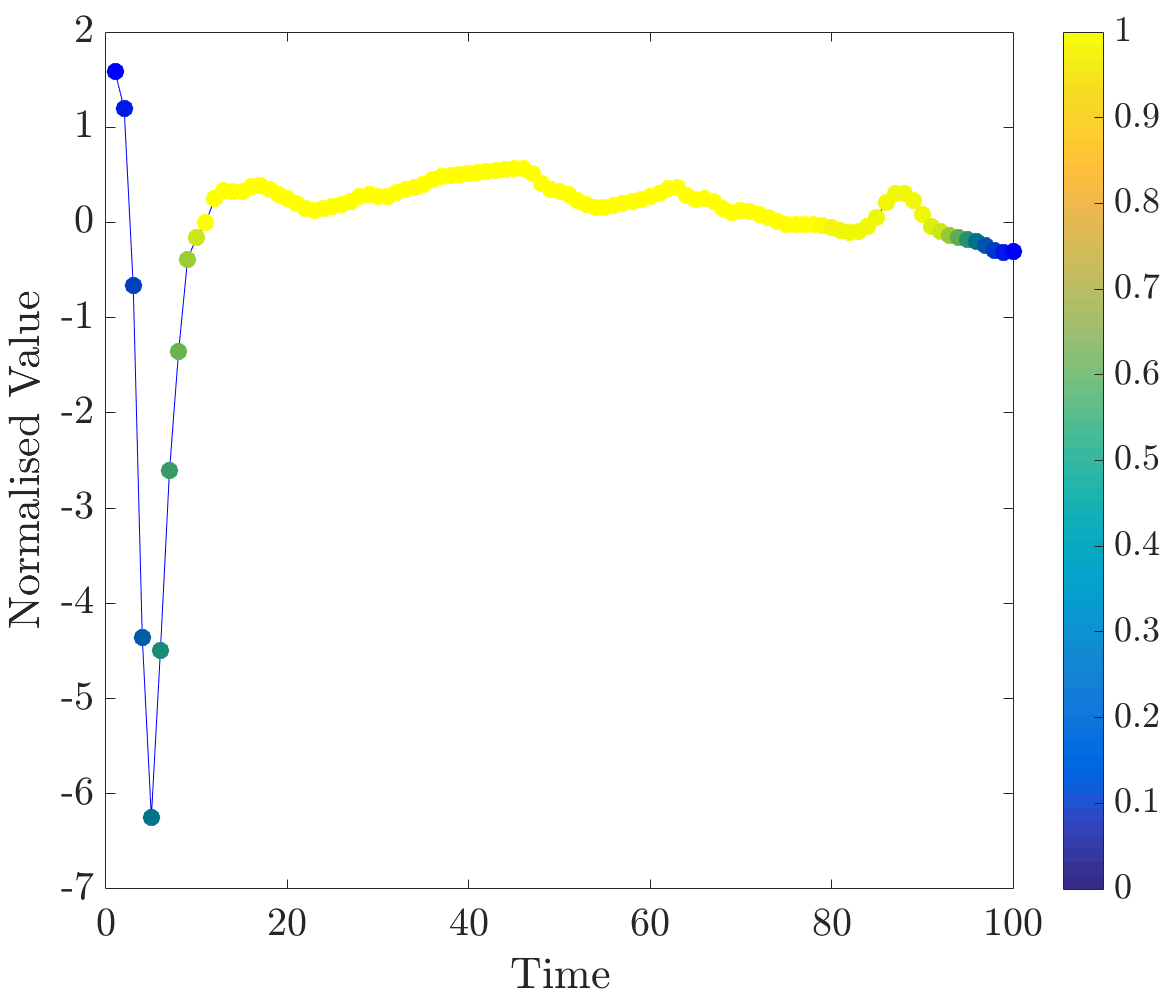}}\quad
  \subfigure[\label{fig:nT}]{\includegraphics[scale=0.11]{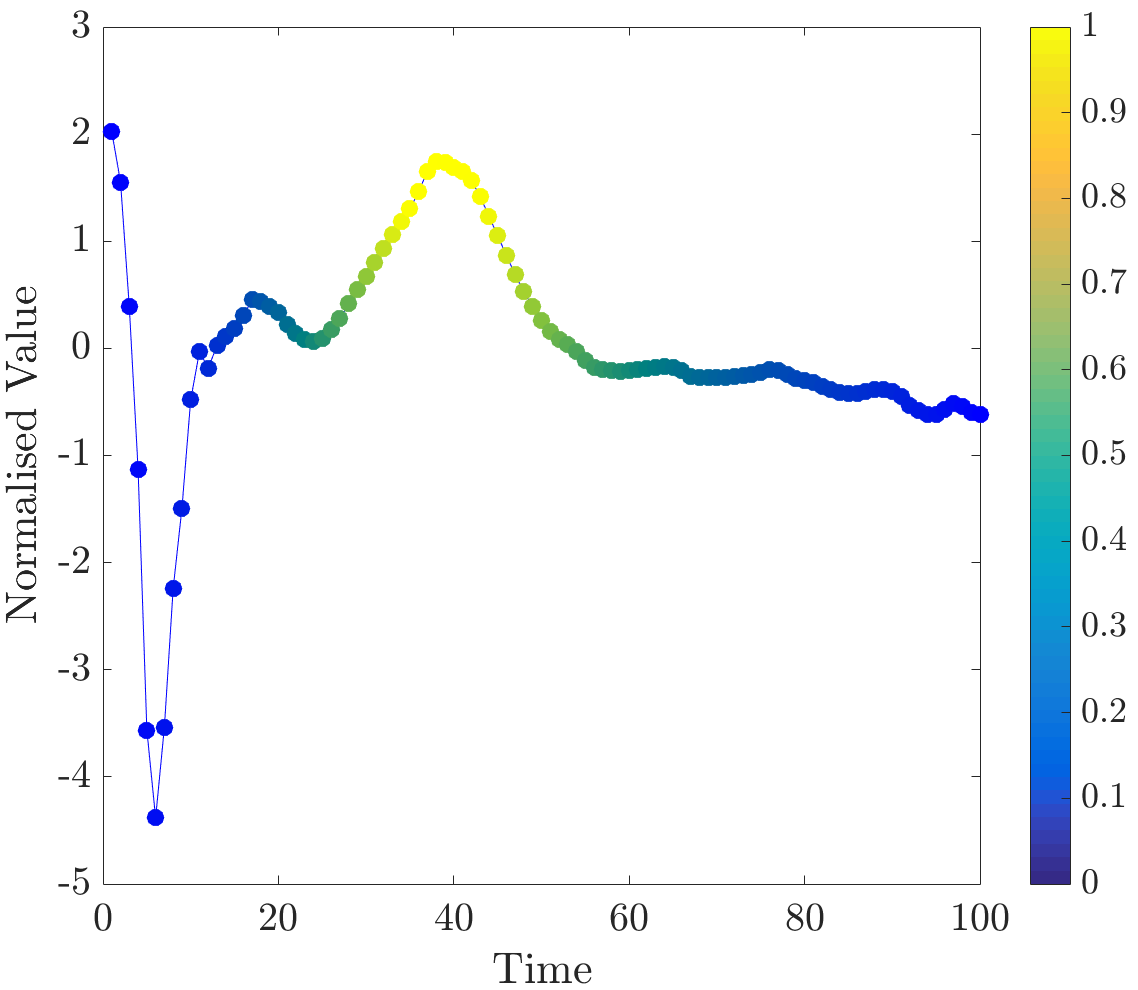}}
  \caption{Normalised relevance values for various beats. Points with high normalised relevance values are plotted in yellow colour. \label{fig:beatsARVD}}
\end{figure}
On the other hand, in almost all normal beats high relevance values were consistently centered around $T$ waves (see Figure \ref{fig:nT}). Nevertheless, there were beats with uninformative distributions of relevance values, an example of such ECG is plotted in Figure \ref{fig:abnStr}. Observe that in the given figures the normalised relevance of a point is shown by its colour, wherein yellow corresponds to high values of the measure and blue stands for low values.

In general, ECG beat segments highlighted by relevance values agree with the segments emphasised by the expert \cite{marcinkevics2017}. In this dataset, highly relevant points are clustered together and their positions agree with our expectations. It is noteworthy that in a few beats, which did not have inverted $T$ waves or high amplitude $P$ waves, relevance is not very useful for discovering important subsequences, for example, this case can be observed in the abnormal time series plotted in Figure \ref{fig:abnStr}. %The limitation of this experiment is a very small number of training instances, however, it allowed for a transparent interpretation of the entire dataset.

\subsubsection{ECG5000}

Another dataset studied is ECG5000, retrieved from the UCR archive. Only two largest classes of beats were kept. Figure \ref{fig:ECG5000data} contains a plot of all 500 ECG beats that were retrieved. The origin of these beats and the biomedical meaning of classes are unknown to us. Therefore, we first provide a description of the differences that can be observed visually.
\begin{figure}[ht]
\begin{center}
\includegraphics[width=6.5cm]{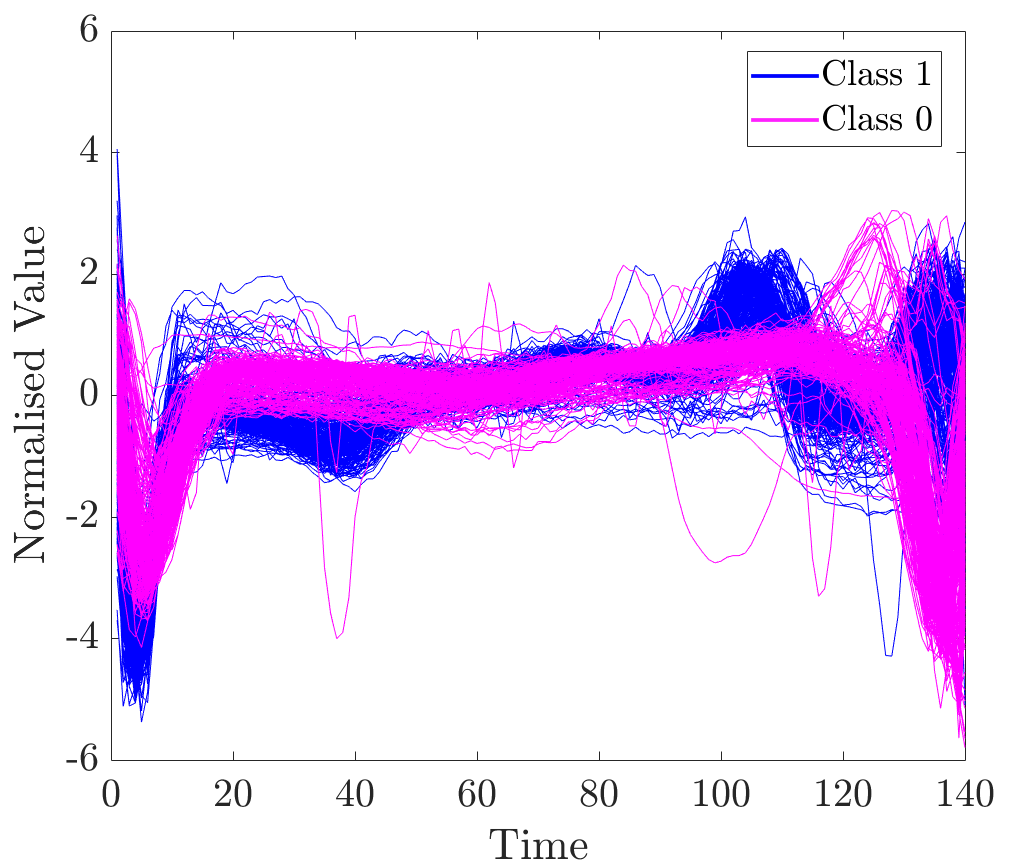}
\caption{ECG beats from ECG5000 dataset.\label{fig:ECG5000data}}
\end{center}
\end{figure}
As can be seen, major differences between classes are, approximately, located at time intervals $[0,20]$ and $[90,140]$. As opposed to class 0, beats of class 1 contain two peaks at times 100 and 135. In this experiment relevance values for several time series are computed and compared to the description derived from the visual inspection.

Several unobserved time series from class 1 were considered \steven{using the same `leave-one-out' tactic described in the ARVD subsection}. In all of these beats large relevance values were assigned to  the points located around, approximately, time 100. This agrees with the timing of the peak that distinguishes many time series of class 1 from class 0. This result is within expectations. Examples of normalised relevance values for several ECG beats can be seen in Figure \ref{fig:beatsECG5000}.
\begin{figure}
  \centering
  \subfigure[\label{fig:ECG5000_1}]{\includegraphics[scale=0.14]{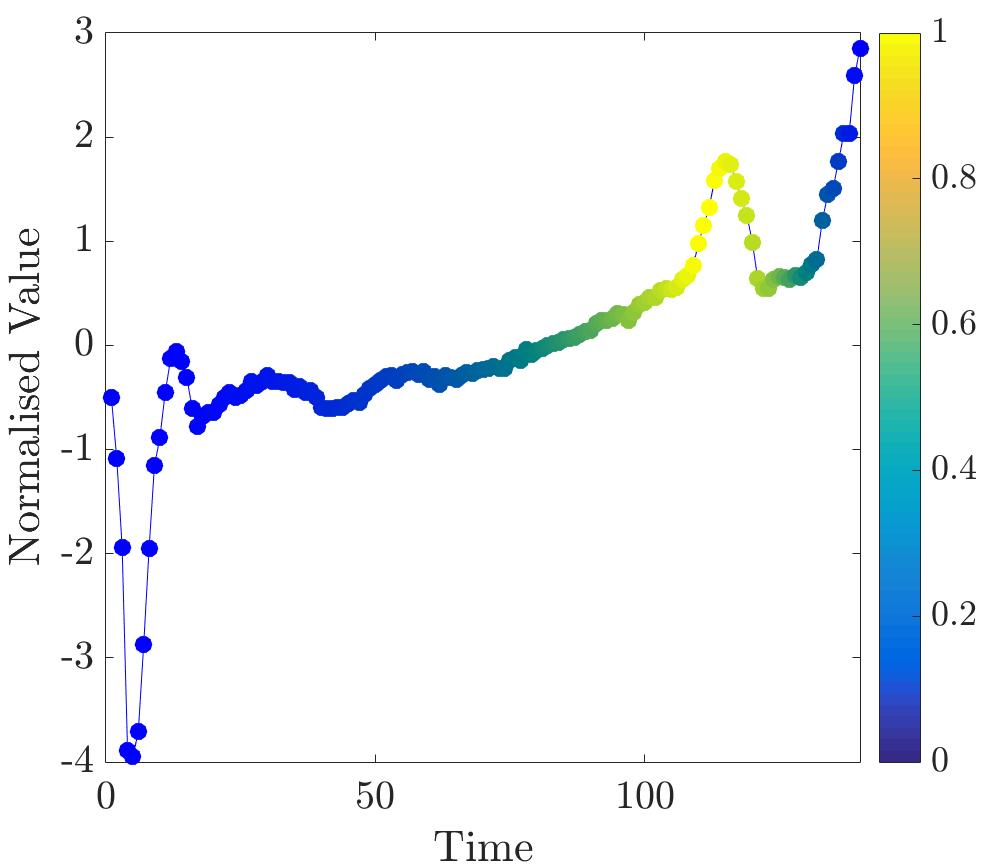}}\quad
  \subfigure[\label{fig:ECG5000_2}]{\includegraphics[scale=0.14]{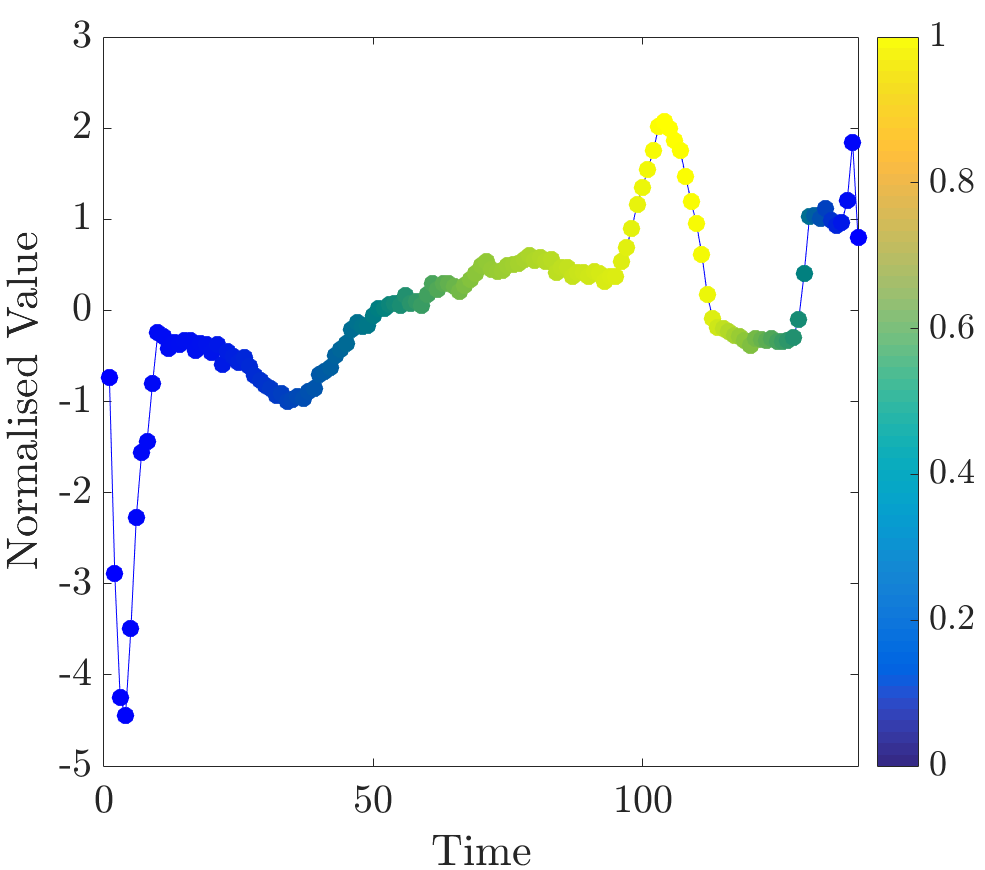}}\quad
  \subfigure[\label{fig:ECG5000_3}]{\includegraphics[scale=0.14]{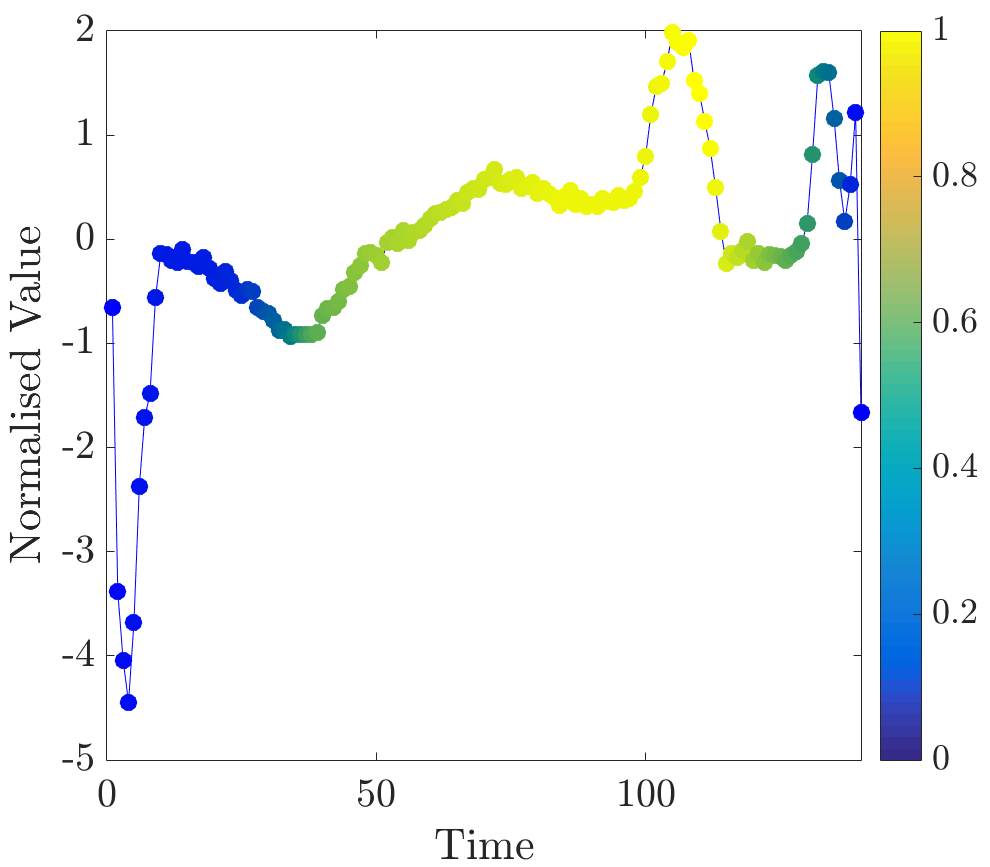}}\quad
  \subfigure[\label{fig:ECG5000_4}]{\includegraphics[scale=0.14]{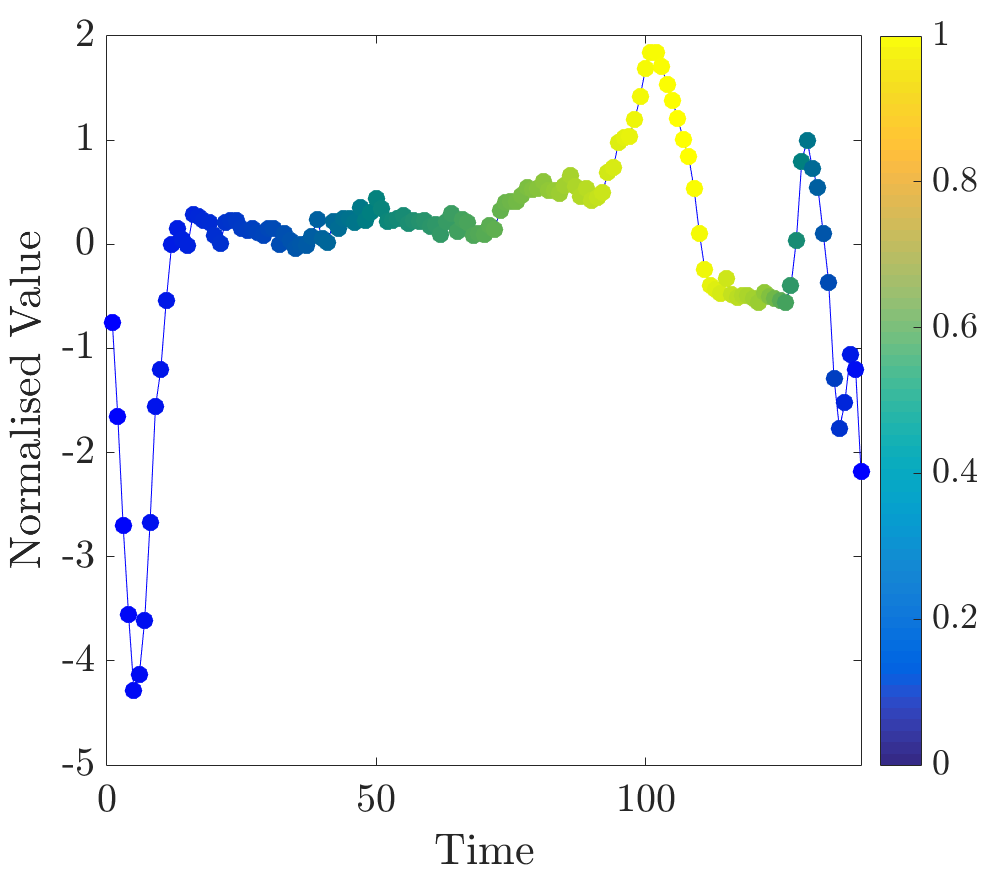}}
  \caption{Normalised relevance values for beats from class 1. Points with high normalised relevance values are plotted in yellow colour.\label{fig:beatsECG5000}}
\end{figure}
In some cases relevant points are quite dense and are restricted to one very short segment (see Figure \ref{fig:ECG5000_1}); on the other hand, sometimes they are dispersed across a longer time interval (as in Figure \ref{fig:ECG5000_3}).

To sum up, in this experiment greater relevance values were assigned to the segments that are distinguishing and noticeable when inspecting beats visually. In this dataset the proposed measure can be used to discover subsequences that are important to the outcome of the classification for class 1.

\subsection{ECG Segment Detection}
\label{subsec:detect}
Another experiment performed is the detection of ECG segments, in particular, the detection of $QRS$ complexes and $T$ waves. Herein we demonstrate how datasets can be constructed such that relevance values can be further used to detect the specified segments in unobserved time series. \steven{Unlike the earlier experiments, here we are using our method in a targetted way to locate types of subsequences that we have identified \emph{a priori} as being of interest.}

Given a set of ECG beats with the end-points of desired segments annotated, a modified labelled dataset has to be constructed. One class is formed by the given time series, and the other class consists of modified time series which result from the removal of labelled segments. This procedure is schematically shown in Figure \ref{fig:construction} for an ECG beat with the labelled $QRS$ complex. \steven{The high-level intuition is that the feature we are looking for needs to be deleted in order to bring the time series close to the class of instances where the feature has been spliced out.}
\begin{figure}[ht]
\begin{center}
\includegraphics[width=6.5cm]{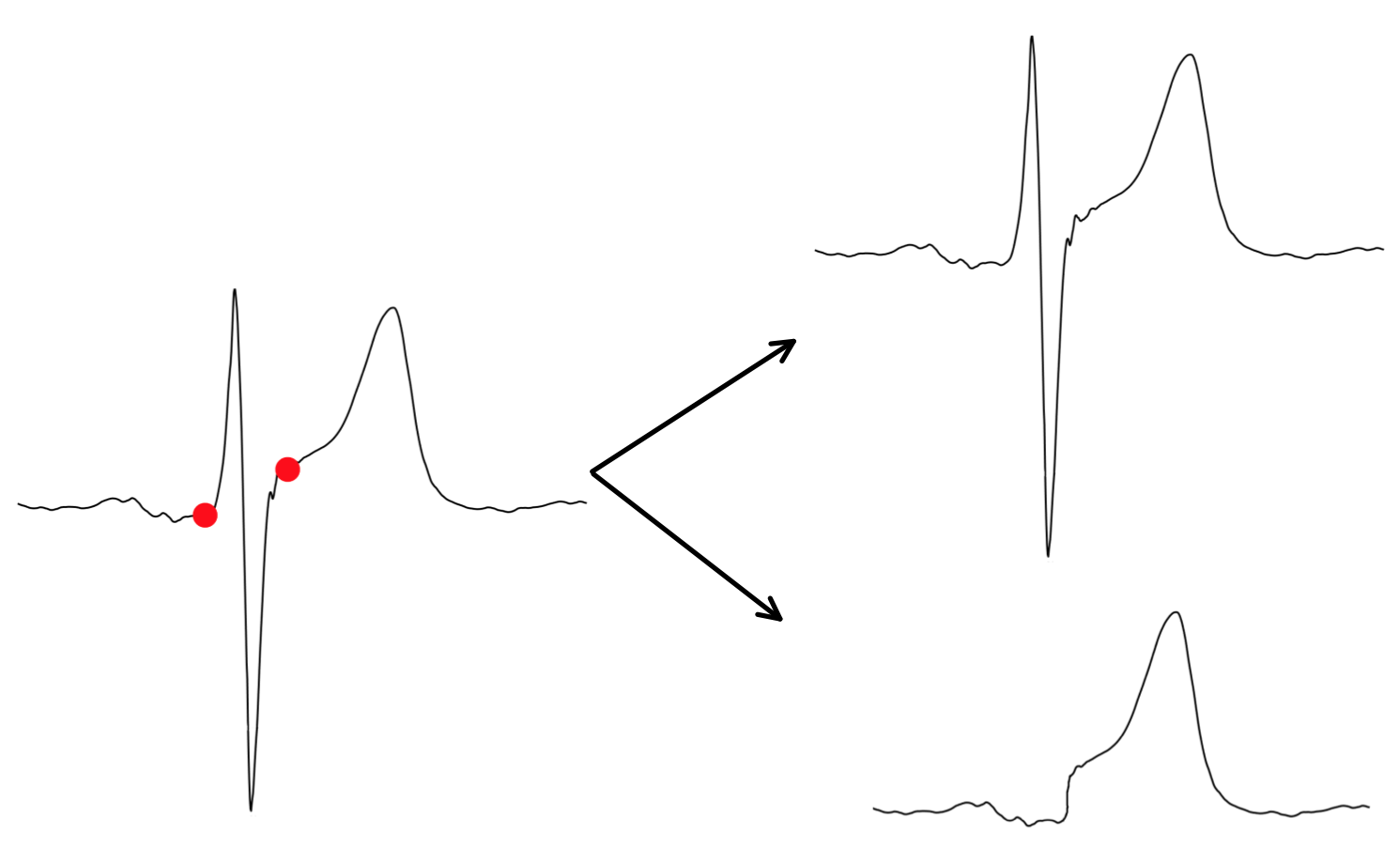}
\caption{Dataset construction for ECG beat segment detection. The dataset contains two classes of ECG beats, with and without the desired segment, respectively. Herein, the $QRS$ complex is labelled in the beat. A new time series is obtained by removing the labelled segment. \label{fig:construction}}
\end{center}
\end{figure}
We demonstrate that datasets constructed in this manner can be exploited in the detection of specified ECG segments. 

\subsubsection{$QRS$ Complex Detection}

Detecting $QRS$ complexes is a fairly standard ECG segmentation task, widely addressed in the literature, for instance, see \cite{sharma2017}. An example of a common approach is the algorithm proposed by Pan and Tompkins in 1985 \cite{pan1985}. In order to perform QRS detection, a set of 50 beats was chosen from CinC\_ECG\_torso, acquired from the UCR archive. All of the ECG beats are plotted in Figure \ref{fig:QRSdata}.
\begin{figure}[ht]
\begin{center}
\includegraphics[width=6.5cm]{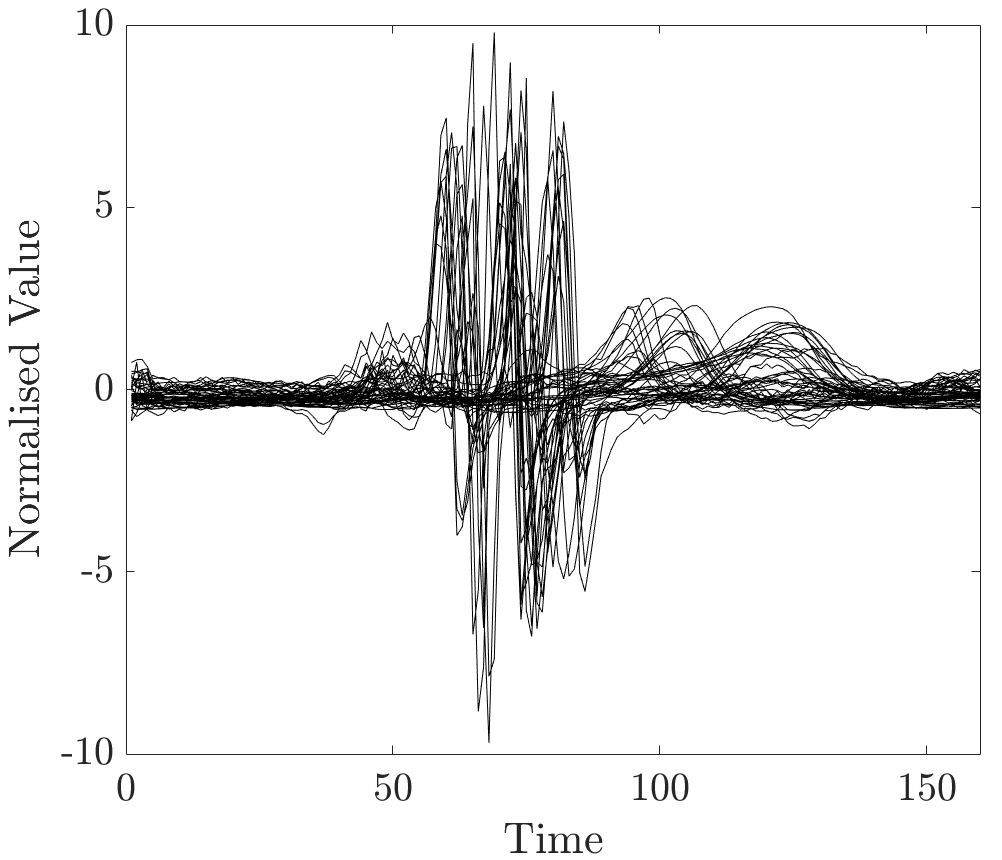}
\caption{ECG beats for $QRS$ complex detection.\label{fig:QRSdata}}
\end{center}
\end{figure}

The modified dataset was constructed as described before by manually labelling $QRS$ segments in the training instances; and relevance values were computed for several unobserved ECG beats. As can be seen from plots in Figure \ref{fig:beatsQRS}, high relevance values clearly align with the locations of $QRS$ complexes in time series. 
\begin{figure}[ht]
  \centering
  \subfigure[\label{fig:qrs1}]{\includegraphics[scale=0.14]{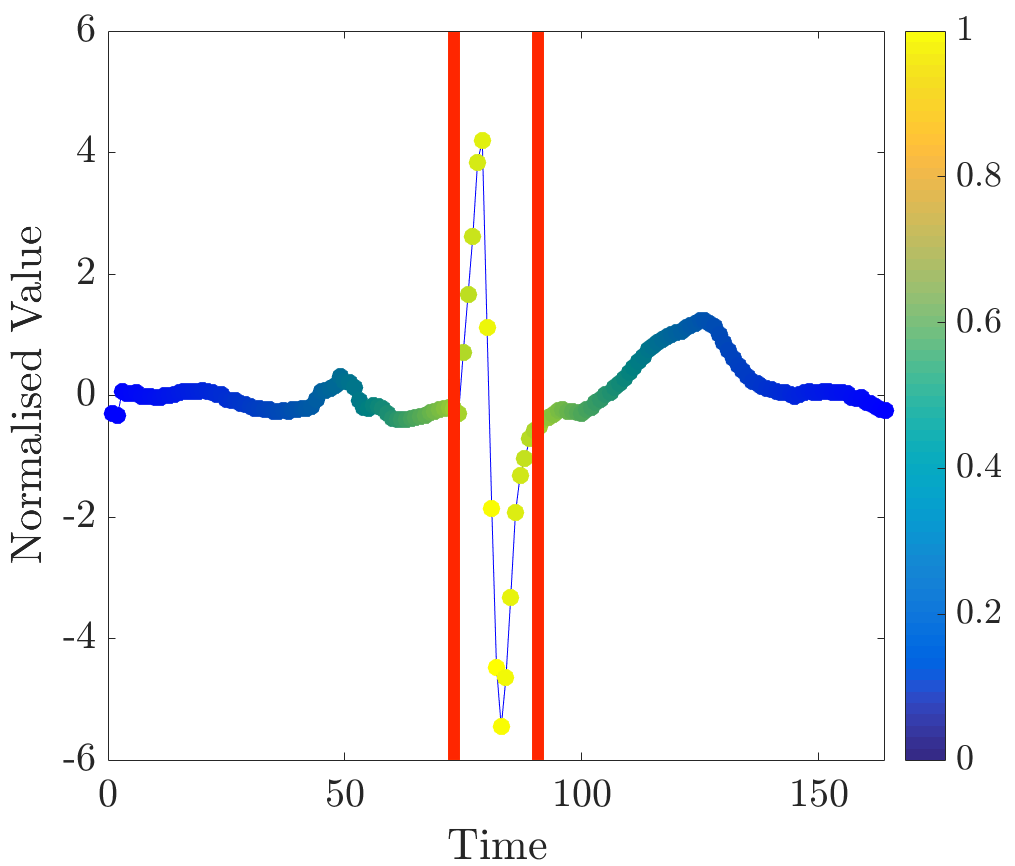}}\quad
  \subfigure[\label{fig:qrs2}]{\includegraphics[scale=0.14]{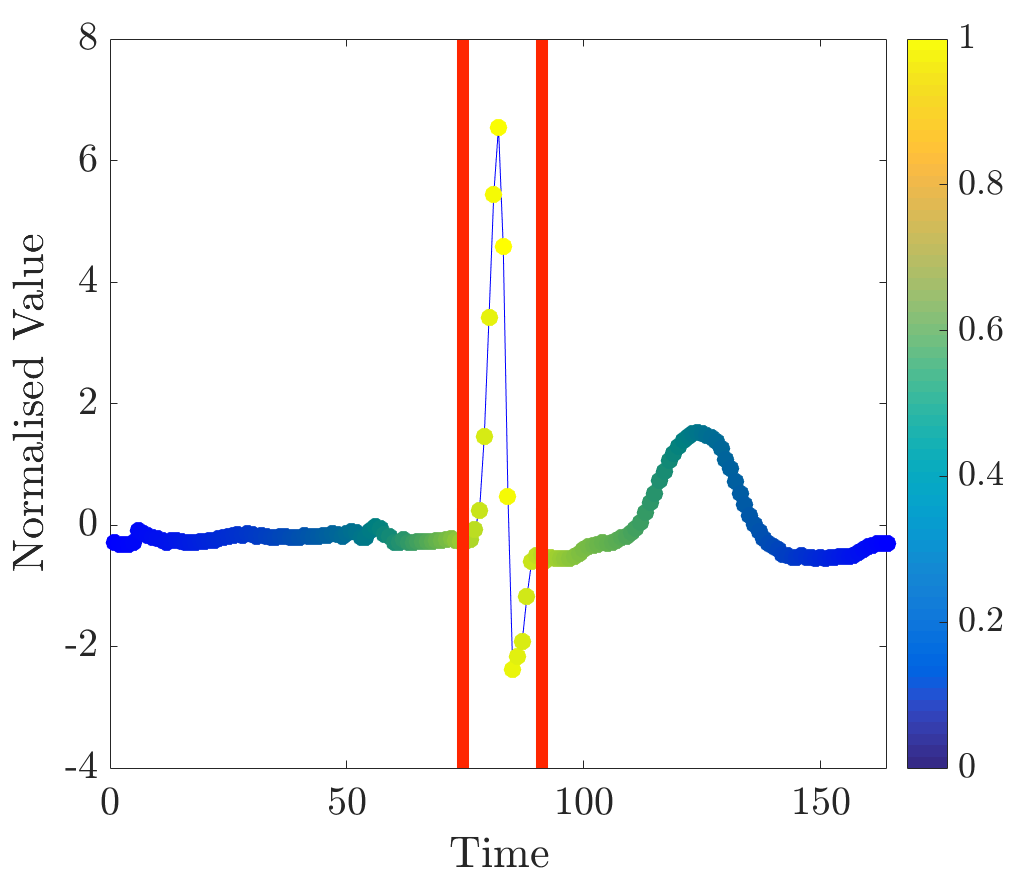}}
  \subfigure[\label{fig:qrs3}]{\includegraphics[scale=0.14]{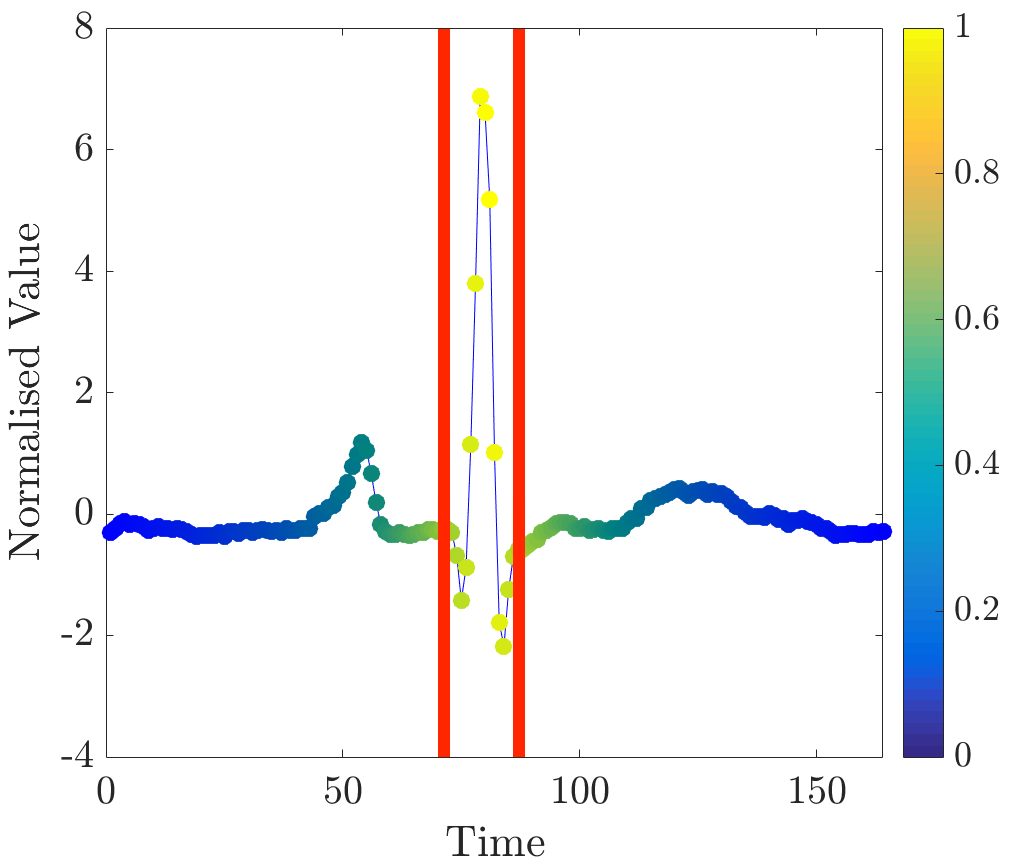}}\quad
  \subfigure[\label{fig:qrs4}]{\includegraphics[scale=0.14]{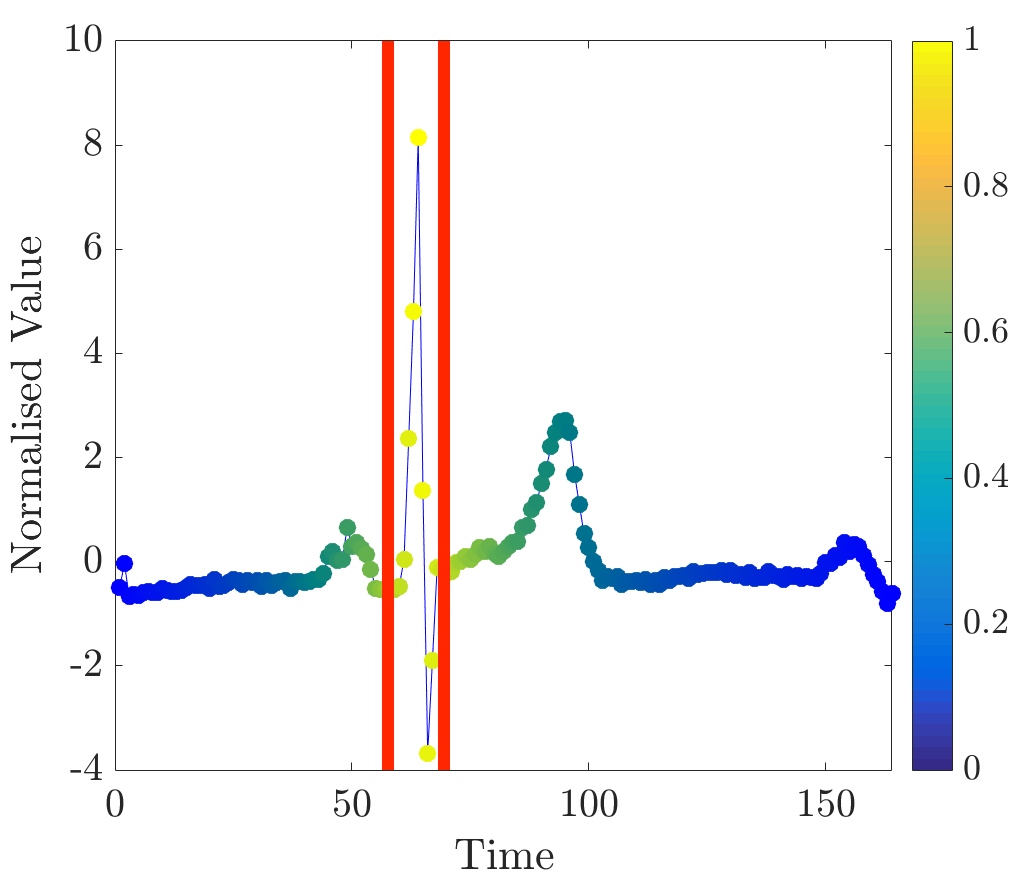}}
  \caption{Normalised relevance values for beats for $QRS$ detection. Points with high normalised relevance values are plotted in yellow colour. End-points of $QRS$ complexes as detected by the expert are shown in red. \label{fig:beatsQRS}}
\end{figure}
Observe that the beats differ in shapes and timings. In addition, ECGs were annotated by an expert; $QRS$ end-points, as detected by the expert, are marked by red lines. The detection of the exact end-point locations was performed by imposing a threshold on the values and retrieving a contiguous segment with points the relevance of which is greater than the threshold (in this case twice the mean relevance was taken as a threshold value). Figure \ref{fig:beatsQRSdet} shows segments as detected by the algorithm; points of segments are marked in yellow.

\begin{figure}[ht]
  \centering
  \subfigure[\label{fig:qrs1det}]{\includegraphics[scale=0.14]{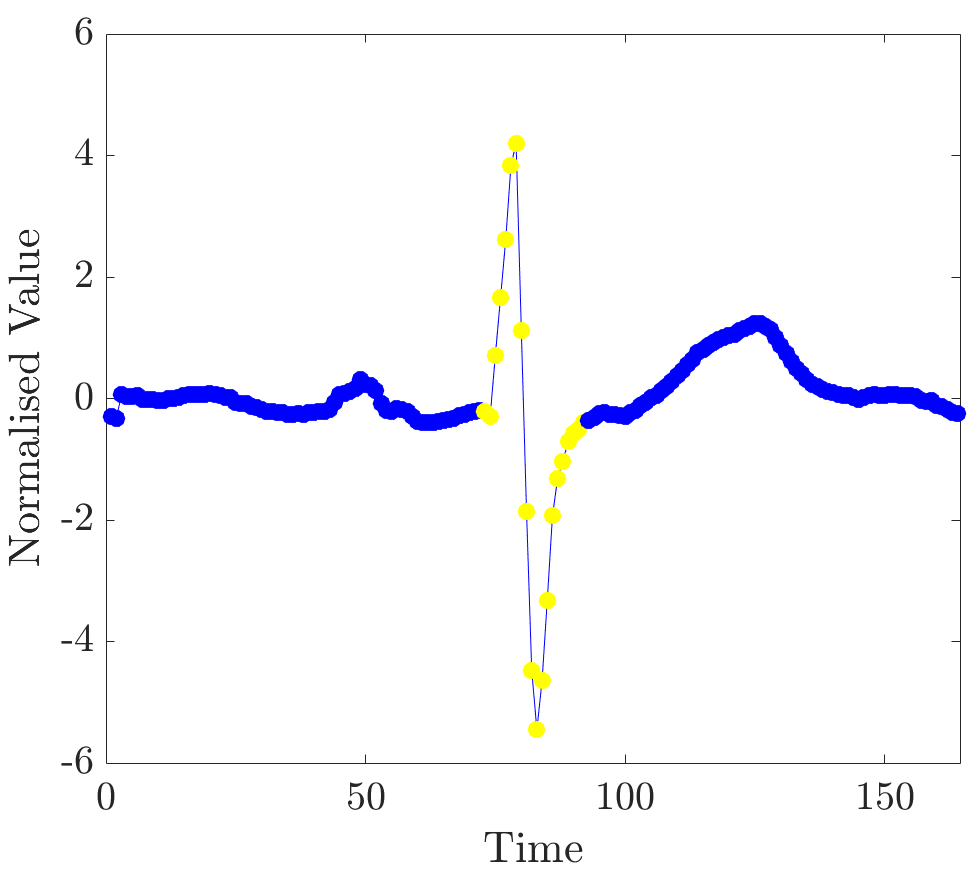}}\quad
  \subfigure[\label{fig:qrs2det}]{\includegraphics[scale=0.14]{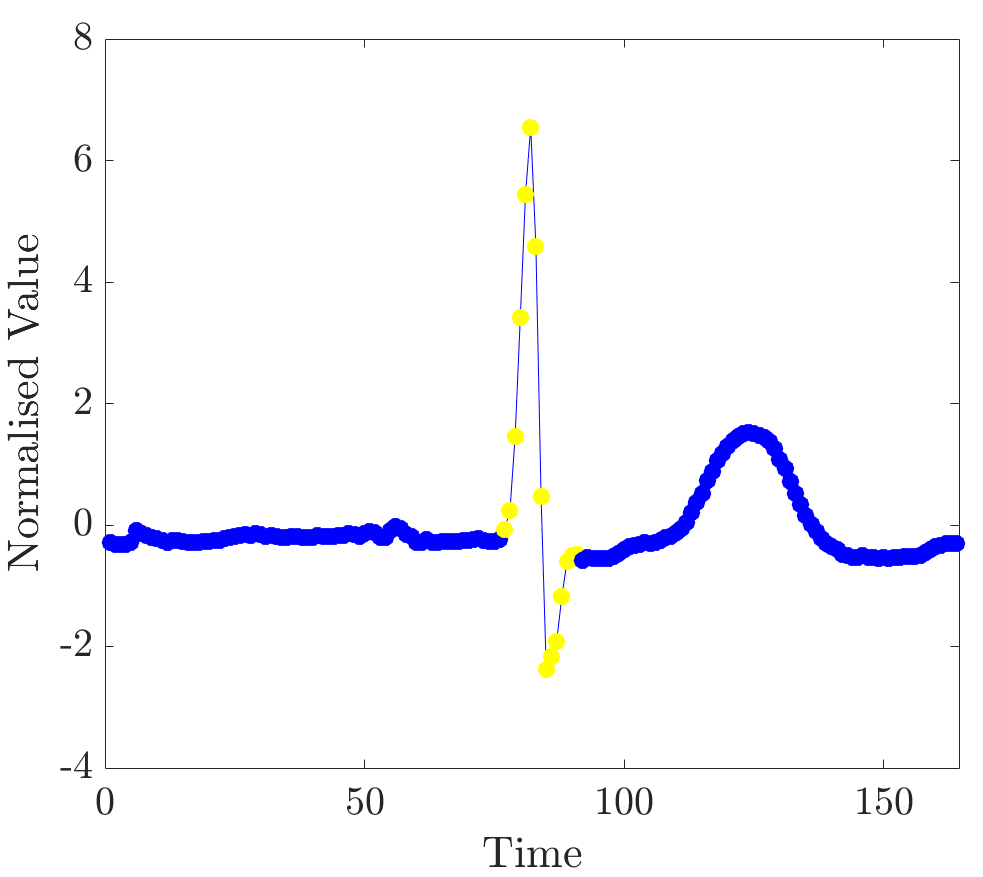}}
  \subfigure[\label{fig:qrs3det}]{\includegraphics[scale=0.14]{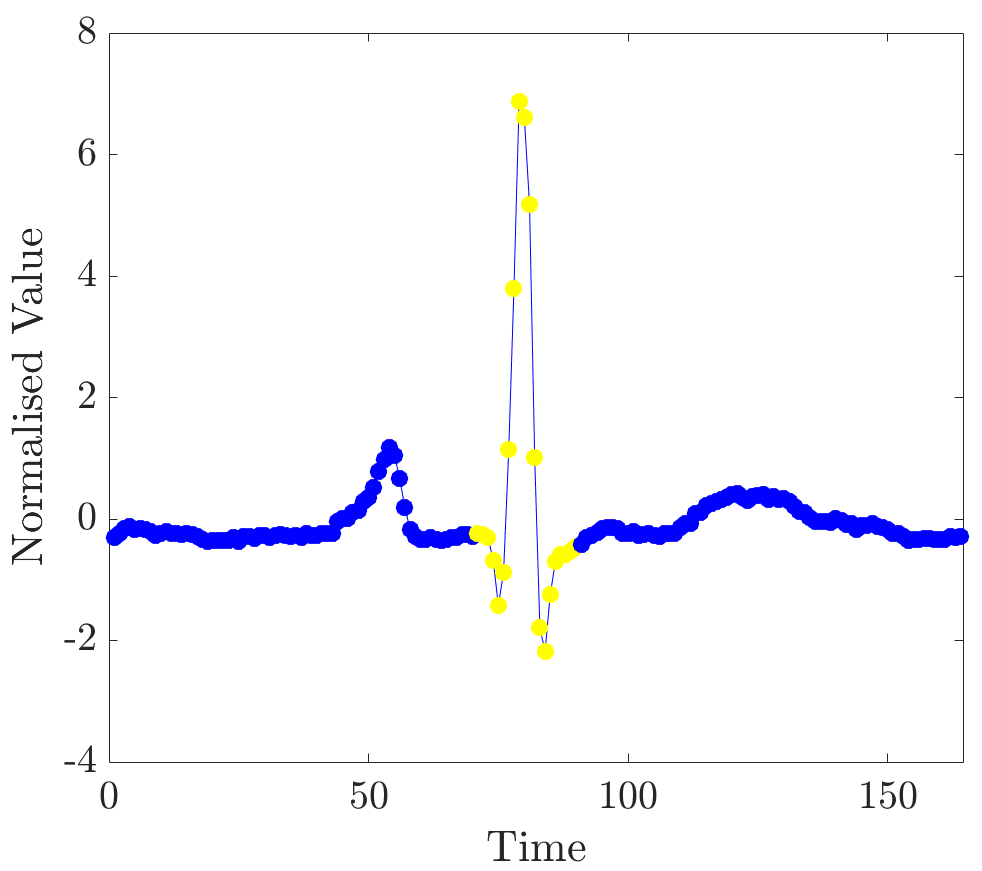}}\quad
  \subfigure[\label{fig:qrs4det}]{\includegraphics[scale=0.14]{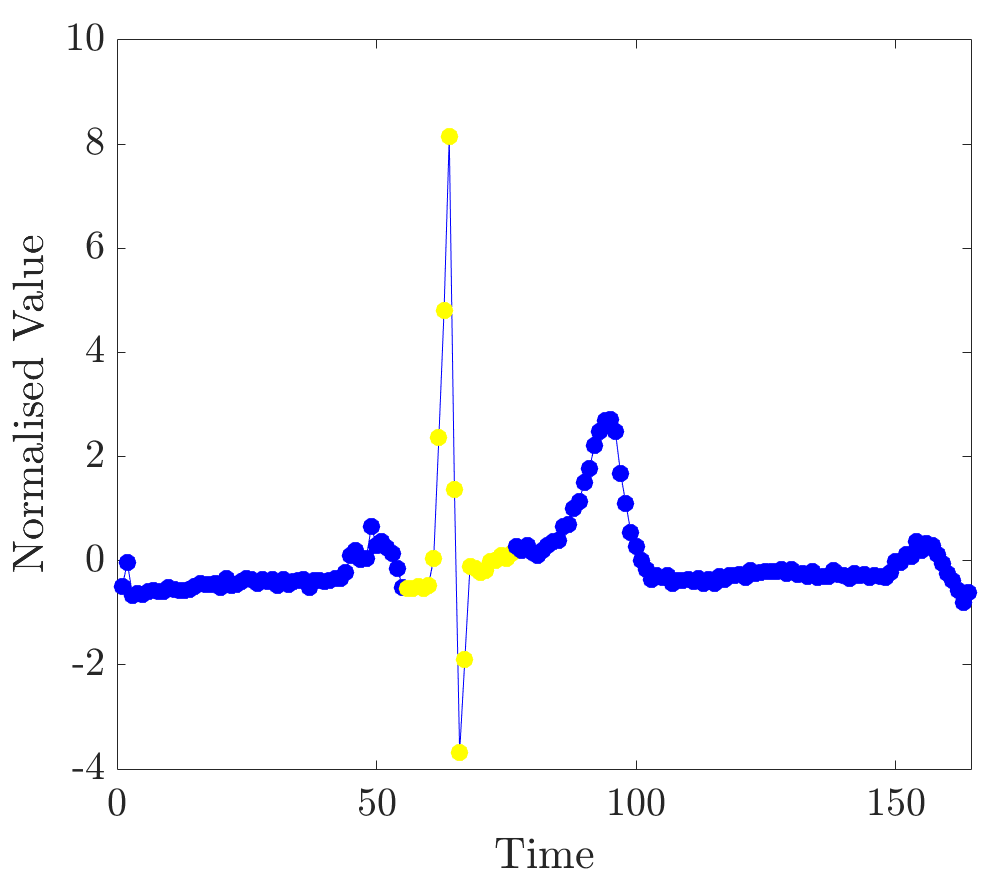}}
  \caption{$QRS$ detection by thresholding on relevance values. Points in $QRS$ complexes as detected by the algorithm are plotted in yellow. \label{fig:beatsQRSdet}}
\end{figure}

In all unobserved instances relevance values robustly indicated QRS complexes. This experiment demonstrates the viability of this approach to the detection of segments at the level of individual ECG beats.

\subsubsection{$T$ Wave Detection}

The detection of $T$ waves is a challenging task, which can be difficult even for human observers \cite{madeiro2013}. For this experiment a dataset was constructed in the manner mentioned above; beats were taken from various sources available at PhysioBank databases. All 50 ECGs used for the construction of the two-class dataset are plotted in Figure \ref{fig:Tdata}.
\begin{figure}[ht]
\begin{center}
\includegraphics[width=6.5cm]{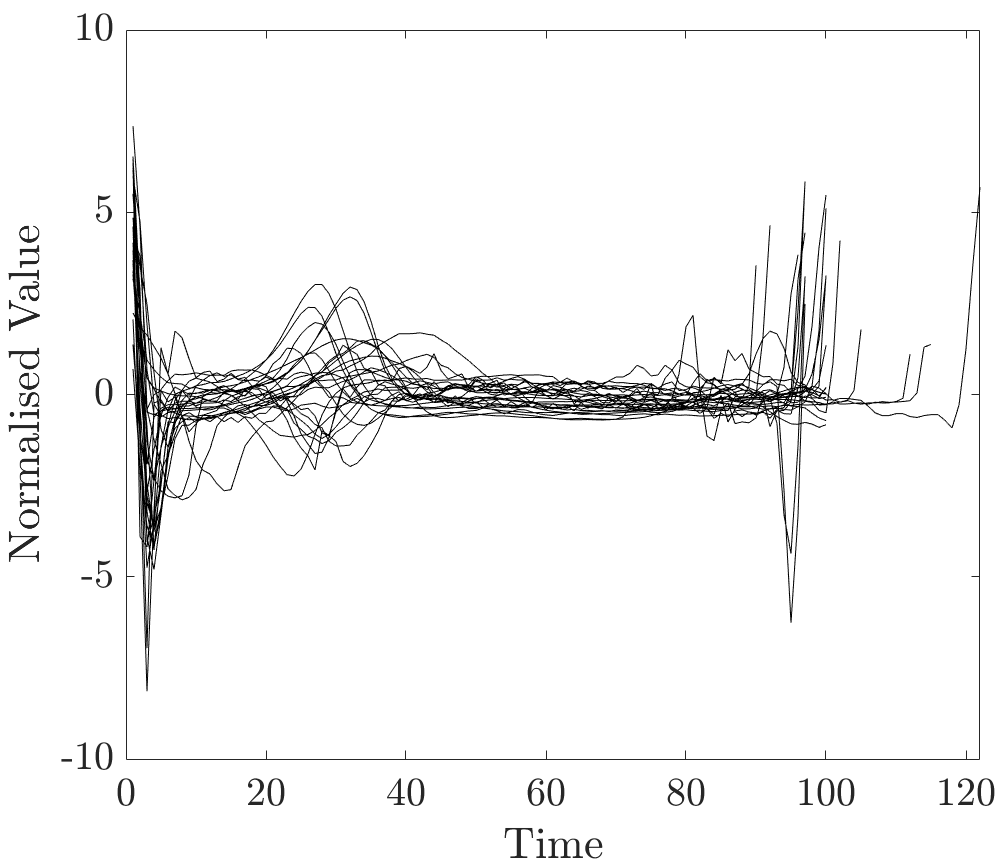}
\caption{ECG beats for $T$ wave detection.\label{fig:Tdata}}
\end{center}
\end{figure}
It is noteworthy that beats have varying lengths and different shapes of $T$ waves, including negative polarity. $T$ waves were manually labelled in all ECGs and the modified set was constructed. Relevance values were computed for several unobserved sequences with various shapes of $T$ waves.

Relevance for various beats is visualised in Figure \ref{fig:beatsTwave}. Red lines mark the end-points of $T$ waves as annotated by the expert. In most of the unobserved ECGs, relevance values are greater for points around $T$ wave peaks.
\begin{figure}
  \centering
  \subfigure[\label{fig:twave_1}]{\includegraphics[scale=0.14]{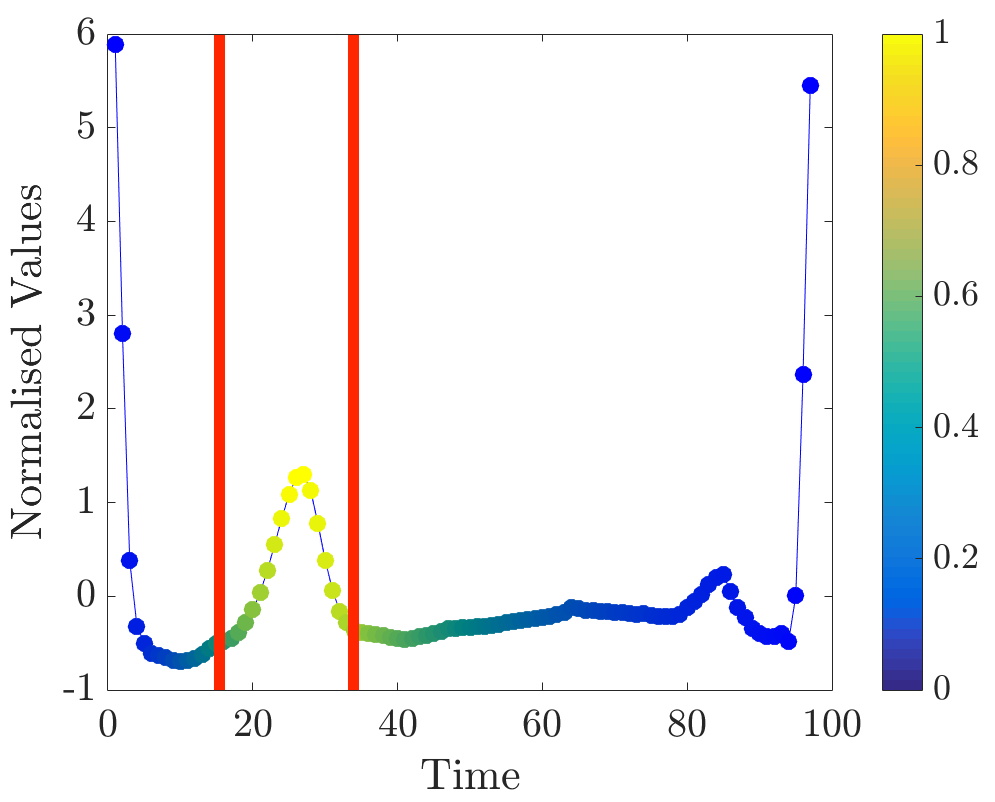}}\quad
  \subfigure[\label{fig:twave_2}]{\includegraphics[scale=0.14]{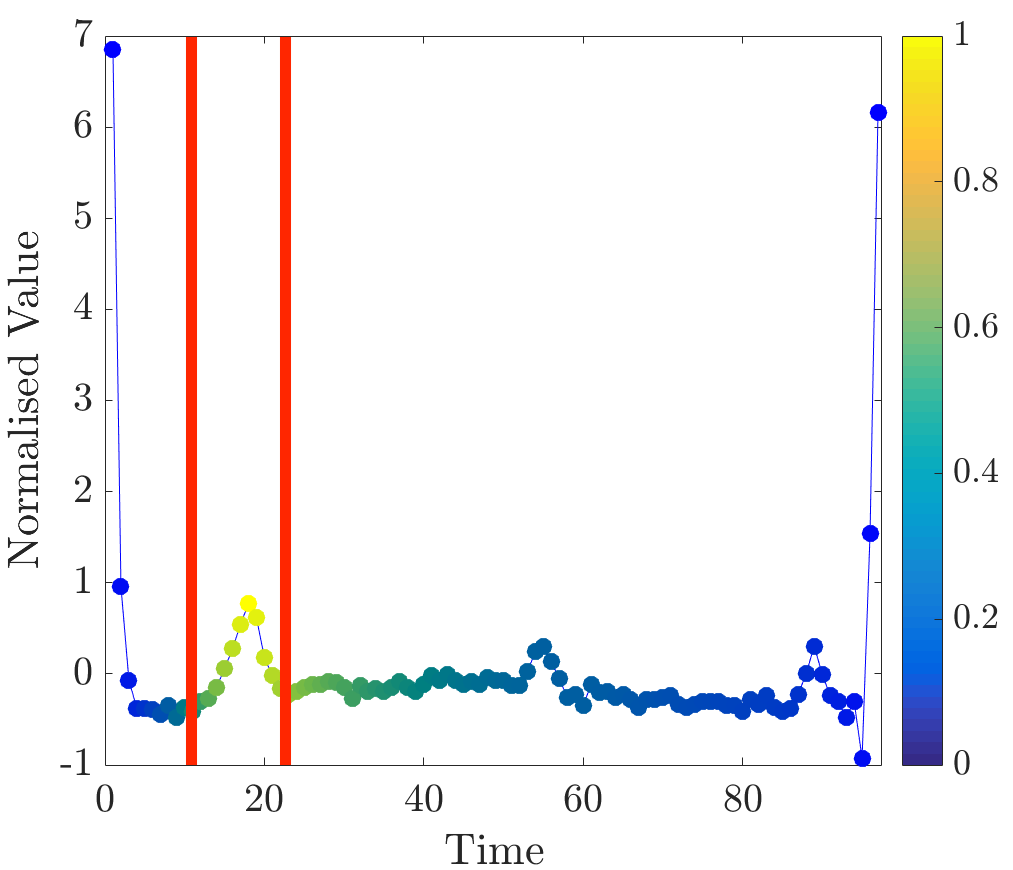}}
  \subfigure[\label{fig:twave_3}]{\includegraphics[scale=0.14]{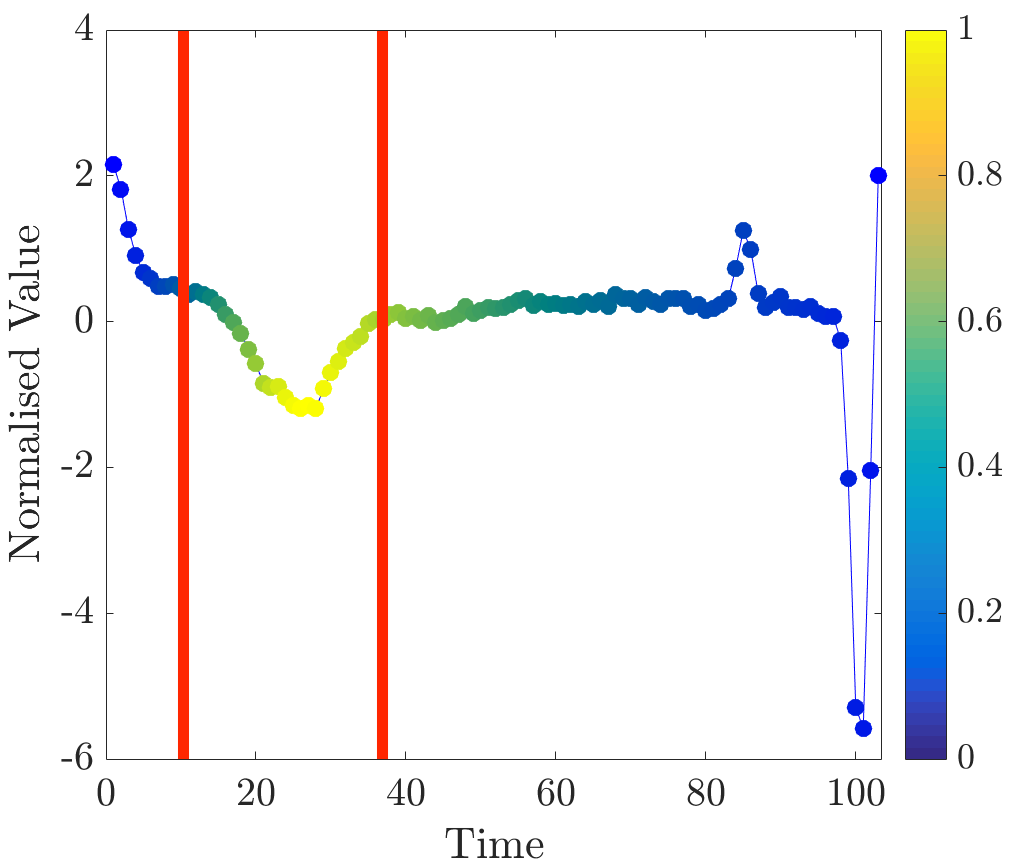}}\quad
  \subfigure[\label{fig:twave_4}]{\includegraphics[scale=0.14]{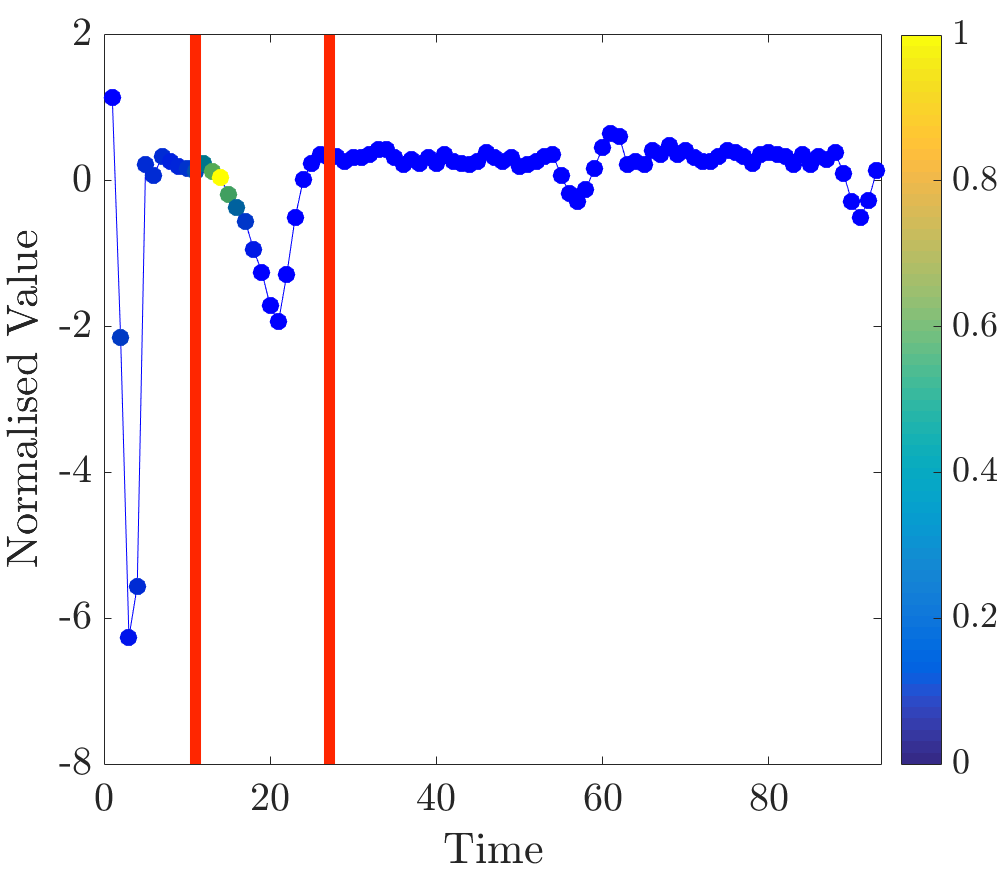}}
  \caption{Normalised relevance values for $T$ wave detection. Points with high normalised relevance values are plotted in yellow colour. End-points of $T$ waves as detected by the expert are shown in red.\label{fig:beatsTwave}}
\end{figure}
In some cases relevance values do not clearly emphasise $T$ wave segments, see Figure \ref{fig:twave_4} for an example wherein inverted $T$ wave could not be detected. The location of end-points can be performed in the same way as for $QRS$ complexes.

In general, in most of the unobserved ECGs, relevance values indicated locations of $T$ waves with different amplitudes and of positive and negative polarities. Nevertheless, in some cases the measure failed to emphasise the segment. This experiment shows that the proposed approach may be applied to the task of $T$ wave location in ECG beats, given a sufficiently representative dataset with labelled segments.

\section{Conclusions} \label{concl.}

We introduced the problem of finding a minimum length subsequence of the given time series the removal of which changes the outcome of classification under the nearest neighbour algorithm with DTW distance. We designed a \steven{simple but optimized} algorithm to find a minimum \textit{contiguous} subsequence; and we also proposed a measure to quantify relevance of every point in the given time series for the considered classification. The described problem and the measure have not been considered in the literature before. We showed how the relevance measure can be used to interpret ECG classification and to perform the detection of segments in ECG beats. Based on the experimental results, it is clear that for considered classification tasks relevance values are able to identify time series segments important for classification outcomes.

Further research can include a more thorough investigation of possible applications of the relevance measure for interpreting classification outcomes of physiological time series. It would be interesting to examine its usage for time series data mining in larger datasets. Additionally, an efficient algorithm needs to be proposed to solve the problem with non-contiguous subsequences. \steven{A natural place to start could be to consider non-contiguous subsequences that consist of only $b$ disjoint contiguous segments, for small $b$.} \steven{Finally, a deeper mathematical understanding of how DTW
distance behaves under the action of subsequence deletion (in the context of the NN algorithm) will hopefully feed the development of more enhanced data structures for our method, further reducing the running time.} 

%$n$\nobreakdash-contiguous subsequences, \textit{i.e.} subsequences consisting of $n$ disjoint contiguous segments, are a particularly noteworthy special case.  Is 

\section*{Acknowledgement}

The authors would like to thank Dr. Muzahir Tayebjee and Dr. Arun Holden for sharing with us ECGs of patients diagnosed with ARVD, and Rachel Cavill for useful discussions.

\newpage

\bibliographystyle{IEEEtran}
\bibliography{ricards}

\begin{thebibliography}{10}
\providecommand{\url}[1]{#1}
\csname url@samestyle\endcsname
\providecommand{\newblock}{\relax}
\providecommand{\bibinfo}[2]{#2}
\providecommand{\BIBentrySTDinterwordspacing}{\spaceskip=0pt\relax}
\providecommand{\BIBentryALTinterwordstretchfactor}{4}
\providecommand{\BIBentryALTinterwordspacing}{\spaceskip=\fontdimen2\font plus
\BIBentryALTinterwordstretchfactor\fontdimen3\font minus
  \fontdimen4\font\relax}
\providecommand{\BIBforeignlanguage}[2]{{%
\expandafter\ifx\csname l@#1\endcsname\relax
\typeout{** WARNING: IEEEtran.bst: No hyphenation pattern has been}%
\typeout{** loaded for the language `#1'. Using the pattern for}%
\typeout{** the default language instead.}%
\else
\language=\csname l@#1\endcsname
\fi
#2}}
\providecommand{\BIBdecl}{\relax}
\BIBdecl

\bibitem{weightedDTW}
{Jeong, Y.-S.}, {Jeong, M. K}., and {Omitaomu, O. A.}, ``Weighted dynamic time
  warping for time series classification,'' \emph{Pattern Recognition},
  vol.~44, no.~9, pp. 2231--2240, Sep. 2011.

\bibitem{elasticEnsemble}
{Lines, J.} and {Bagnall, A.}, ``Time series classification with ensembles of
  elastic distance measures,'' \emph{Data Mining and Knowledge Discovery},
  vol.~29, no.~3, pp. 565--592, May 2015.

\bibitem{periodogram}
{Caiado, J.}, {Crato, N.}, and {Pe\~{n}a, D.}, ``A periodogram-based metric for
  time series classification,'' \emph{Computational Statistics and Data
  Analysis}, vol.~50, no.~10, pp. 2668--2684, Jun. 2006.

\bibitem{svm}
{Rodr\'{\i}guez, J. J.}, {Alonso, C. J.}, and {Maestro, J. A.}, ``Support
  vector machines of interval-based features for time series classification,''
  \emph{Knowledge-Based Systems}, vol.~18, no. 4-5, pp. 171--178, Aug. 2005.

\bibitem{rnn}
{H\"{u}sken, M.} and {Stagge, P.}, ``Recurrent neural networks for time series
  classification,'' \emph{Neurocomputing}, vol.~50, pp. 223--235, 2003.

\bibitem{berndt1994}
{Berndt, D.J.} and {Clifford, J.}, ``Using dynamic time warping to find
  patterns in time series,'' in \emph{AAAIWS'94 Proceedings of the 3rd
  International Conference on Knowledge Discovery and Data Mining}, August
  1994, pp. 359--370.

\bibitem{bagnall2017}
{Bagnall, A.}, {Lines, J.}, {Bostrom, A.}, {Large, J.}, and {Keogh, E.}, ``The
  great time series classification bake off: a review and experimental
  evaluation of recent algorithmic advances,'' \emph{Data Mining and Knowledge
  Discovery}, vol.~31, no.~3, pp. 606--660, 2017.

\bibitem{goodman2016european}
{Goodman, B.} and {Flaxman, S.}, ``European union regulations on algorithmic
  decision-making and a ``right to explanation'','' \emph{arXiv preprint
  arXiv:1606.08813}, 2016.

\bibitem{marcinkevics2017}
{Marcinkevics, R.}, {O'Neill, J.}, {Law, H.}, {Pervolaraki, E.}, {Hogarth, A.},
  {Russell, C.}, {Stegemann, B.}, {Holden, A.V.}, and {Tayebjee, M.H.},
  ``Multichannel {ECG} diagnostics for the diagnosis of arrhythmogenic right
  ventricular dysplasia,'' \emph{EP Europace}, vol. eux124, 2017.

\bibitem{ding2008}
{Ding, H.}, {Trajcevski, G.}, {Scheuermann, P.}, {Wang, X.}, and {Keogh, E.},
  ``Querying and mining of time series data: Experimental comparison of
  representations and distance measures,'' in \emph{Proceedings of the VLDB
  Endowment}, vol.~1, August 2008, pp. 1542--1552.

\bibitem{esling2012}
{Esling, P.} and {Agon, P.}, ``Time-series data mining,'' \emph{{A}{C}{M}
  Computing Surveys}, vol.~45, no.~1, pp. 12--34, 2012.

\bibitem{CoverHart1967}
{Cover, T.} and {Hart, P.}, ``Nearest neighbor pattern classification,''
  \emph{IEEE Transactions on Information Theory}, vol.~13, no.~1, pp. 21--27,
  Sep. 2006.

\bibitem{wilson1972}
{Wilson, D. L.}, ``Asymptotic properties of nearest neighbor rules using edited
  data,'' \emph{IEEE Transactions on Systems, Man, and Cybernetics}, vol.
  SMC-2, no.~3, pp. 408--421, July 1972.

\bibitem{Friedman1975}
{Friedman, J. H.}, {Baskett, F.}, and {Shustek, L. J.}, ``An algorithm for
  finding nearest neighbors,'' \emph{IEEE Transactions on Computers}, vol.~24,
  no.~10, pp. 1000--1006, Oct. 1975.

\bibitem{mitchell1997}
{Mitchell, T.M.}, ``Instance-based learning,'' in \emph{Machine Learning},
  1st~ed.\hskip 1em plus 0.5em minus 0.4em\relax McGraw-Hill, 1997, ch.~8, pp.
  230--248.

\bibitem{clarkson2006}
{Clarkson, K.L.}, ``Nearest-neighbor searching and metric space dimensions,''
  in \emph{Nearest-Neighbor Methods for Learning and Vision: Theory and
  Practice}.\hskip 1em plus 0.5em minus 0.4em\relax MIT Press, 2006.

\bibitem{pauleve2010}
{Paulev\'{e}, L.}, {J\'{e}gou, H.}, and {Amsaleg, L.}, ``Locality sensitive
  hashing: a comparison of hash function types and querying mechanisms,''
  \emph{Pattern Recognition Letters}, vol.~31, no.~11, pp. 1348--1358, 2010.

\bibitem{rakthanmanon2012}
{Rakthanmanon, T.}, {Campana, B.}, {Mueen, A.}, {Batista, G.}, {Westover, B.},
  {Zhu, Q.}, {Zakaria, J.}, and {Keogh, E.}, ``Searching and mining trillions
  of time series subsequences under dynamic time warping,'' in
  \emph{Proceedings of the 18th ACM SIGKDD International Conference on
  Knowledge Discovery and Data Mining}, 2012, pp. 262--270.

\bibitem{kim2001}
{Kim, S.}, {Park, S.}, and {Chu, W.W.}, ``An index-based approach for
  similarity search supporting time warping in large sequence databases,'' in
  \emph{Proceedings of the 17th International Conference on Data Engineering},
  2001, pp. 607--614.

\bibitem{ahmed2017}
{Ahmed, R.}, {Temko, A.}, {Marnane, W.P.}, {Boylan, G.}, and {Lightbody, G.},
  ``Exploring temporal information in neonatal seizures using a dynamic time
  warping based {S}{V}{M} kernel,'' \emph{Computers in Biology and Medicine},
  vol.~82, pp. 100--110, 2017.

\bibitem{muller2007}
{M{\"u}ller, M.}, ``Dynamic time warping,'' in \emph{Information Retrieval for
  Music and Motion}.\hskip 1em plus 0.5em minus 0.4em\relax Springer-Verlag
  Berlin Heidelberg, 2007, pp. 69--84.

\bibitem{keogh2005}
{Keogh, E.} and {Ratanamahatana, C.A.}, ``Exact indexing of dynamic time
  warping,'' \emph{Knowledge and Information Systems}, vol.~7, no.~3, pp.
  358--386, 2005.

\bibitem{yi1998}
{Yi, B.-K.}, {Jagadish, H.V.}, and {Faloutsos, C.}, ``Efficient retrieval of
  similar time sequences under time warping,'' in \emph{Proceedings of the 14th
  International Conference on Data Engineering}, 1998.

\bibitem{salvador2007}
{Salvador, S.} and {Chan, P.}, ``Toward accurate dynamic time warping in linear
  time and space,'' \emph{Intelligent Data Analysis}, vol.~11, no.~5, pp.
  561--580, 2007.

\bibitem{karel2009}
{Karel, J.M.H.}, ``Cardiac signal processing,'' in \emph{A wavelet approach to
  cardiac signal processing for low-power hardware applications}, 2009.

\bibitem{Susto2018}
{Susto, G. A.}, {Cenedese, A.}, and {Terzi, M.}, ``Chapter 9 - time-series
  classification methods: Review and applications to power systems data,'' in
  \emph{Big Data Application in Power Systems}, {Arghandeh, R.} and {Zhou, Y.},
  Eds.\hskip 1em plus 0.5em minus 0.4em\relax Elsevier, 2018, pp. 179 -- 220.

\bibitem{keogh2005disc}
{Keogh, E.}, {Lin, J.}, and {Fu, A.}, ``{H}{O}{T} {S}{A}{X}: Efficiently
  finding the most unusual time series subsequence,'' in \emph{Proceedings of
  the 5th IEEE International Conference on Data Mining}, 2005, pp. 226--233.

\bibitem{ye2009}
{Ye, L.} and {Keogh, E.}, ``Time series shapelets: a new primitive for data
  mining,'' in \emph{Proceedings of the 15th ACM SIGKDD International
  Conference on Knowledge Discovery and Data Mining}, 2009, pp. 947--956.

\bibitem{silva2016}
{Silva, D.F.} and {Batista, G.E.A.P.A.}, ``Speeding up all-pairwise dynamic
  time warping matrix calculation,'' in \emph{Proceedings of the 2016 SIAM
  International Conference on Data Mining}, 2016, pp. 837--845.

\bibitem{ucrarchive}
{Chen, Y.}, {Keogh, E.}, {Hu, B.}, {Begum, N.}, {Bagnall, A.}, {Mueen, A.}, and
  {Batista, G.}, ``The {U}{C}{R} time series classification archive,''
  \url{http://www.cs.ucr.edu/~eamonn/time_series_data/}, 2015.

\bibitem{physionet}
{Goldberger, A.L.}, {Amaral, L.A.N.}, {Glass, L.}, {Hausdorff, J.M.}, {Ivanov,
  P.Ch.}, {Mark, R.G.}, {Mietus, J.E.}, {Moody, G.B.}, {Peng, C.-K.}, and
  {Stanley, H.E.}, ``{PhysioBank, PhysioToolkit, and PhysioNet}: Components of
  a new research resource for complex physiologic signals,''
  \emph{Circulation}, vol. 101, no.~23, pp. e215--e220, 2000.

\bibitem{marcus1982}
{Marcus, F.I.}, {Fontaine, G.H.}, {Guiraudon, G.}, {Frank, R.}, {Laurenceau,
  J.L.}, {Malergue, C.}, and {Grosgogeat, Y.}, ``Right ventricular displasia: A
  report of 24 adult cases,'' \emph{Circulation}, vol.~65, no.~2, pp. 384--398,
  1982.

\bibitem{sharma2017}
{Sharma, T.} and {Sharma, K.K.}, ``{Q}{R}{S} complex detection in {E}{C}{G}
  signals using locally adaptive weighted total variation denoising,''
  \emph{Computers in Biology and Medicine}, vol.~87, pp. 187--199, 2017.

\bibitem{pan1985}
{Pan, J.} and {Tompkins, W.J.}, ``A real-time {QRS} detection algorithm,''
  \emph{IEEE Transactions on Biomedical Engineering}, vol. BME-32, no.~3, pp.
  230--236, 1985.

\bibitem{madeiro2013}
{Madeiro, J.P.V.}, {Nicolson, W.B.}, {Cortez, P.C.}, {Marques, J.A.L.},
  {V\'{a}zquez-Seisdedos, C.R.}, {Elangovan, N.}, {Ng, G.A.}, and {Schlindwein,
  F.S.}, ``New approach for {T}-wave peak detection and {T}-wave end location
  in 12-lead paced {ECG} signals based on a mathematical model,'' \emph{Medical
  Engineering and Physics}, vol.~35, no.~8, pp. 1105--1115, 2013.

\end{thebibliography}

\clearpage

\begin{IEEEbiography}[{\includegraphics[width=1in]{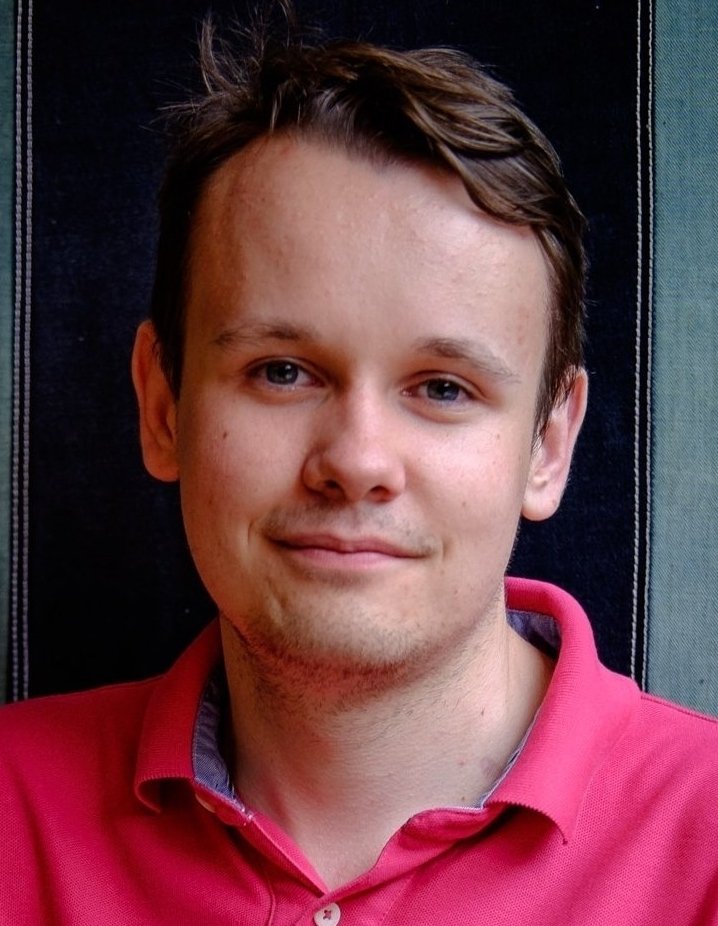}}]{Ri\v{c}ards Marcinkevi\v{c}s}
received the BSc degree in data science and knowledge engineering from Maastricht University, in 2017. From 2015 to 2017, he interned at the Medtronic Bakken Research Center, Maastricht. He worked on biomedical data analysis and signal processing. Currently, he is working towards the MSc degree in statistics at ETH Z\"urich.
\end{IEEEbiography}

\begin{IEEEbiography}[{\includegraphics[width=1in]{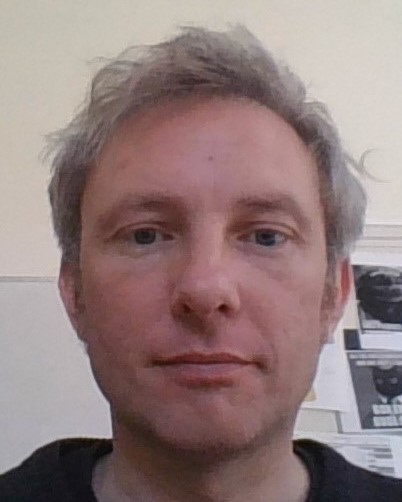}}]{Steven Kelk}
%\begin{IEEEbiographynophoto}{Steven Kelk}
%Steven M. Kelk
is an associate professor
at the Department of Data Science and Knowledge Engineering
(DKE) at Maastricht University in The Netherlands,
where he works on applications of combinatorial
optimization and discrete mathematics
to computational biology. His main research
interests are phylogenetics, graph theory, fixed parameter tractability,
approximation algorithms and (integer) linear programming.
\end{IEEEbiography}

\begin{IEEEbiography}[{\includegraphics[width=1in]{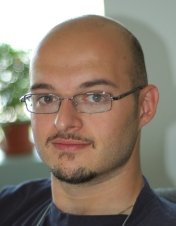}}]{Carlo Galuzzi}
received a M.Sc. degree (cum laude) in mathematics from the Department of Mathematics at the University  of Milan, Italy, in 2003, and a Ph.D. degree in computer engineering from Delft University of Technology in 2009. He was post-doctoral researcher at the same university
from 2009 till 2015. Since 2015, he is assistant professor at Maastricht University, Maastricht, The Netherlands.  His research interests include instruction-set architecture customizations,
reconfigurable and parallel computing, brain-machine interfaces, brain modeling, hardware/software co-
design, mathematical modeling, graph theory, and design space exploration.
\end{IEEEbiography}

%\begin{IEEEbiographynophoto}{Berthold Stegemann}

\begin{IEEEbiography}[{\includegraphics[width=1in]{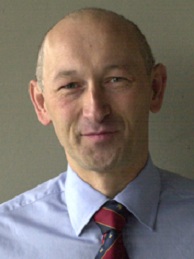}}]{Berthold Stegemann}
%Berthold Stegemann
is a Senior Principal Scientist at the Research \& Technology department of the Bakken Research Center (BRC) in Maastricht, Netherlands. He does applied research on implantable electronic devices to improve implantable electronic device functions. His main interests are medical data analysis, and algorithm development.
\end{IEEEbiography}

%\end{biography}

\end{document}